\documentclass[conference]{IEEEtran}
\IEEEoverridecommandlockouts
\usepackage{cite}
\usepackage{amsmath,amssymb,amsfonts,bm}
\usepackage[ruled]{algorithm2e}
\usepackage{booktabs}
\usepackage{algpseudocode}
\usepackage{graphicx}
\usepackage{multirow}
\usepackage{subcaption}
\usepackage{textcomp}
\usepackage{xcolor}
\usepackage{enumitem,kantlipsum}
\usepackage{colortbl}
\definecolor{cwmgroup}{RGB}{255, 243, 220}      % warm amber tint (light)
\definecolor{cwmgroupdark}{RGB}{210, 140, 30}   % warm amber dark
\definecolor{dpcgroup}{RGB}{220, 235, 255}      % cool blue tint (light)
\definecolor{dpcgroupdark}{RGB}{125, 180, 255}   % cool blue dark
\usepackage[table,dvipsnames]{xcolor}
\usepackage{xfp}
\usepackage{etoolbox}
\usepackage{tikz}
\usepackage{pgfplots}
\usepackage{pgfplotstable}
\pgfplotsset{compat=1.18}
\usepgfplotslibrary{groupplots}
\pgfplotsset{compat=newest}

\newcommand{\gradientcell}[5]{%
  \edef\percent{\fpeval{round((#1 - #2) / (#3 - #2) * 100, 0)}}%
  \expanded{\noexpand\cellcolor{#4!\percent!white}}%
  #1\ifblank{#5}{}{{\scriptsize\ (#5)}}% Delta in scriptsize for better hierarchy
}
\usepackage{mdframed}

\def\BibTeX{{\rm B\kern-.05em{\sc i\kern-.025em b}\kern-.08em
    T\kern-.1667em\lower.7ex\hbox{E}\kern-.125emX}}
\begin{document}

\title{Context-specific Credibility-aware Multimodal Fusion with Conditional Probabilistic Circuits
% *\\
% {\footnotesize \textsuperscript{*}Note: Sub-titles are not captured for https://ieeexplore.ieee.org  and
% should not be used}
% \thanks{Identify applicable funding agency here. If none, delete this.}
}
\author{
    \IEEEauthorblockN{Pranuthi Tenali\IEEEauthorrefmark{2}\IEEEauthorrefmark{1}, Sahil Sidheekh\IEEEauthorrefmark{2}\IEEEauthorrefmark{1}, Saurabh Mathur\IEEEauthorrefmark{4}\IEEEauthorrefmark{1}\thanks{* - Equal contribution }, Erik Blasch\IEEEauthorrefmark{3}, Kristian Kersting\IEEEauthorrefmark{4}, Sriraam Natarajan\IEEEauthorrefmark{2}}
    \IEEEauthorblockA{\IEEEauthorrefmark{2}The University of Texas at Dallas
    \\\{Sahil.Sidheekh, Pranuthi.Tenali, Sriraam.Natarajan\}@utdallas.edu}
    \IEEEauthorblockA{\IEEEauthorrefmark{4}Department of Computer Science , TU Darmstadt
    \\saurabh.mathur@tu-darmstadt.de, kersting@cs.tu-darmstadt.de}
    \IEEEauthorblockA{\IEEEauthorrefmark{3}Air Force Research Lab
    \\erik.blasch.1@us.af.mil}

}
% \author{\IEEEauthorblockN{1\textsuperscript{st} Given Name Surname}
% \IEEEauthorblockA{\textit{dept. name of organization (of Aff.)} \\
% \textit{name of organization (of Aff.)}\\
% City, Country \\
% email address or ORCID}
% \and
% \IEEEauthorblockN{2\textsuperscript{nd} Given Name Surname}
% \IEEEauthorblockA{\textit{dept. name of organization (of Aff.)} \\
% \textit{name of organization (of Aff.)}\\
% City, Country \\
% email address or ORCID}
% \and
% \IEEEauthorblockN{3\textsuperscript{rd} Given Name Surname}
% \IEEEauthorblockA{\textit{dept. name of organization (of Aff.)} \\
% \textit{name of organization (of Aff.)}\\
% City, Country \\
% email address or ORCID}
% \and
% \IEEEauthorblockN{4\textsuperscript{th} Given Name Surname}
% \IEEEauthorblockA{\textit{dept. name of organization (of Aff.)} \\
% \textit{name of organization (of Aff.)}\\
% City, Country \\
% email address or ORCID}
% \and
% \IEEEauthorblockN{5\textsuperscript{th} Given Name Surname}
% \IEEEauthorblockA{\textit{dept. name of organization (of Aff.)} \\
% \textit{name of organization (of Aff.)}\\
% City, Country \\
% email address or ORCID}
% \and
% \IEEEauthorblockN{6\textsuperscript{th} Given Name Surname}
% \IEEEauthorblockA{\textit{dept. name of organization (of Aff.)} \\
% \textit{name of organization (of Aff.)}\\
% City, Country \\
% email address or ORCID}
% }

\maketitle
% CoCMF, CrediFuse, C2MF Context-specific Credibility-aware Multimodal Fusion
% MAC^2 Fusion: Multimodal Adaptive Context-specific Credibility-Aware Fusion
\begin{abstract}
Multimodal fusion requires integrating information from multiple sources that may conflict depending on context. Existing fusion approaches typically rely on static assumptions about source reliability, limiting their ability to resolve conflicts when a modality becomes unreliable due to situational factors such as sensor degradation or class-specific corruption.
We introduce \textbf{C$^2$MF}, a context-specfic credibility-aware multimodal fusion framework that models per-instance source reliability using a Conditional Probabilistic Circuit (CPC). We formalize instance-level reliability through \emph{Context-Specific Information Credibility} (CSIC), a KL-divergence--based measure computed exactly from the CPC. CSIC generalizes conventional static credibility estimates as a special case, enabling principled and adaptive reliability assessment.
To evaluate robustness under cross-modal conflicts, we propose the \textbf{\textit{Conflict}} benchmark, in which class-specific corruptions deliberately induce discrepancies between different modalities. Experimental results show that C$^2$MF improves predictive accuracy by up to 29\% over static-reliability baselines in high-noise settings, while preserving the interpretability advantages of probabilistic circuit-based fusion.

  % Multimodal fusion systems often reconcile conflicting information by modeling source reliabilities. Existing fusion methods often rely on static reliability assumptions, leaving them unable to resolve conflicts when a modality becomes unreliable due to situational factors.
  % We introduce a context-aware, reliable multimodal fusion framework $\text{C}^2\text{MF}$ that dynamically models source reliability based on environment context using a conditional probabilistic circuit (CPC). Our framework uses a joint multimodal embedding to condition the CPC, which is used to fuse unimodal predictions. To evaluate this, we introduce the Conflict benchmark where class-specific corruptions create intentional, context-dependent conflicts between audio and visual sources. Our empirical results demonstrate that our approach achieves an 18\% improvement in predictive accuracy over static reliability baselines while retaining the explainability inherent to PC-based fusion.
\end{abstract}


\section{Introduction}

Decision making in real-world domains requires fusing information from diverse modalities of available data for robust and reliable reasoning \cite{baltruvsaitis2018multimodal, blasch2012high}. In safety-critical domains such as autonomous navigation, industrial robotics, and medical decision support, the redundancy provided by multiple sensors is expected to mitigate the failure of any single modality.
% , as can be the case in domains like autonomous navigation and medical decision-support
However, the presence of multiple sources can also 
% create a situation where 
introduce a fundamental challenge in determining which modality to trust, when
the sources provide \textbf{conflicting information}. 
\begin{figure}[!t]
    \centering
    \includegraphics[width=\linewidth]{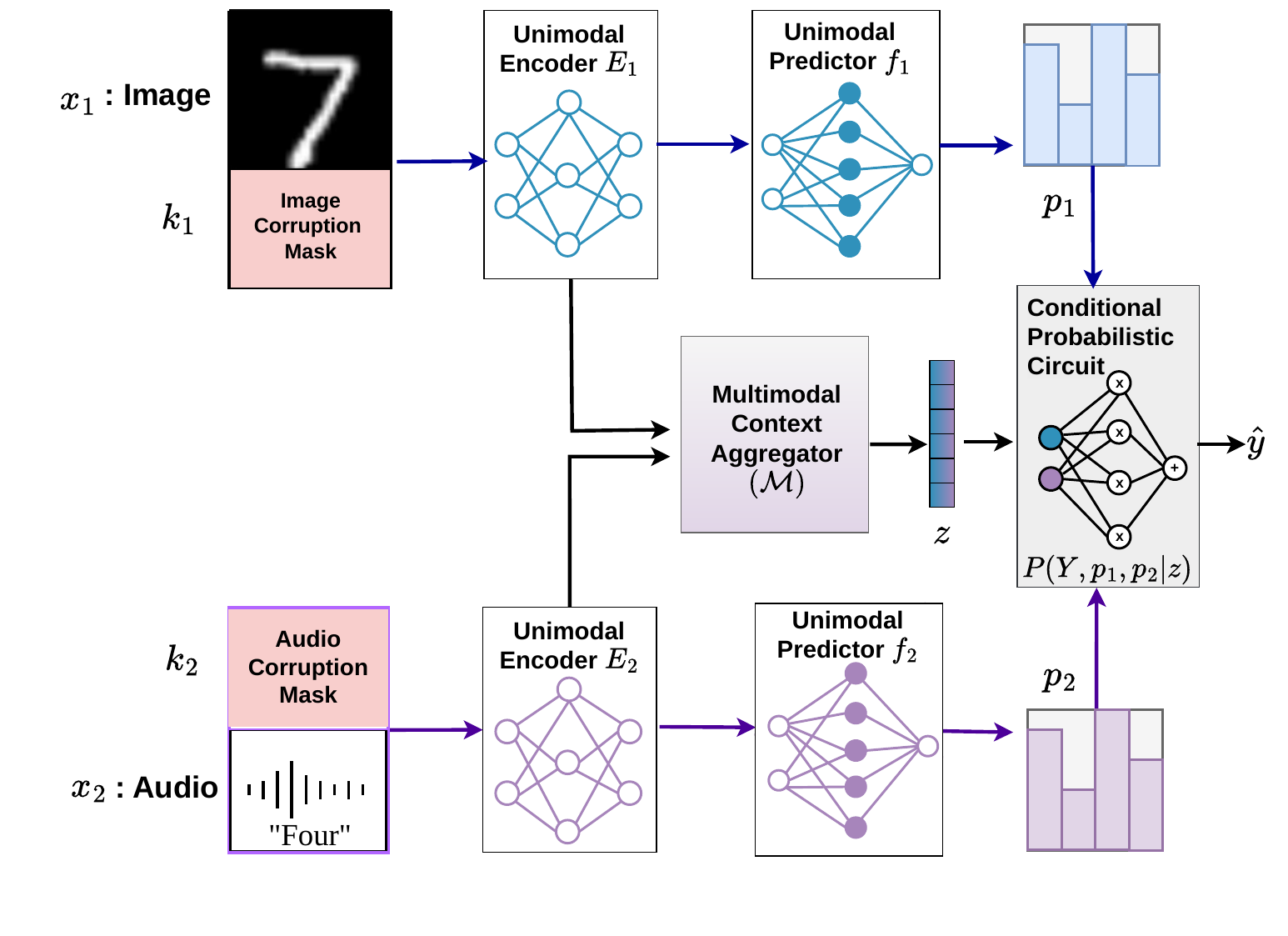}
    \caption{\textbf{Credibility-Aware Fusion using Latent Context}. Unlike static fusion methods, the C$^2$MF framework extracts a joint neural embedding that serves as a context for a Conditional Probabilistic Circuit. The circuit dynamically evaluates the credibility of each source based on its position in the latent space (i.e., identifying when one modality is unreliable), ensuring that the final prediction is dominated by the most reliable information source for that specific instance. }
    \label{fig:c2mf}
    \vspace{-0.2in}
\end{figure}
% Existing fusion architectures typically address such conflicts using meta-information about each source’s reliability.
A common strategy for resolving such conflicts is to model meta-information about each source's reliability \cite{rogova2004reliability, blasch2013urref}.
In such fusion methods, each modality is assigned a weight reflecting the expected credibility or trustworthiness of its information, and the final predictions are aggregated accordingly.
% quantify source reliability as an aggregate of the expected credibility of its information on a training set \cite{sidheekh2025credibility}. 
These reliability estimates are typically learned from data and treated as global properties of the modality.
While effective under stationary conditions, this global assumption breaks down in real-world settings where the reliability of a modality is inherently context-dependent \cite{blasch2014urref, snidaro2016context}. %\cite{}
% However, reliability is rarely stationary in the real world.
For example, a high-resolution camera may be highly reliable in daylight but quite untrustworthy in a low-light tunnel. Similarly, a microphone may provide clear signals in a quiet room but fail in a crowded street. When two highly confident sources disagree, a static reliability model has no principled way to resolve the conflict, increasing the risk of errors.
This observation motivates the central question of our work: \emph{How can a fusion system dynamically infer the context-specific credibility of a modality at an instance level, while retaining principled probabilistic reasoning and semantics?}
Addressing this question effectively
% in high-stakes settings where source reliability varies with context
requires building a multimodal fusion system that satisfies the following desiderata:

\begin{itemize}
    \item[\bf (D1)] It must infer the context directly from the input.
    \item[\bf (D2)] It must dynamically adapt source reliability on an instance-by-instance basis, based on the inferred context.
    \item[\bf (D3)] It must remain tractable, providing exact and explainable reasons for its decisions.
\end{itemize}

% \begin{figure}[t]
%     \centering
%     % \includegraphics[width=.98\linewidth]{Figures/Fusion.pdf}
%     \includegraphics[width=\linewidth]{Figures/c2mf-compact.pdf}
%     \caption{\textbf{Credibility-Aware Fusion using Latent Context}. Unlike static fusion methods, the C$^2$MF framework extracts a joint neural embedding that serves as a context for a Conditional Probabilistic Circuit. The circuit dynamically evaluates the credibility of each source based on its position in the latent space (i.e., identifying when one modality is unreliable), ensuring that the final prediction is dominated by the most reliable information source for that specific instance. }
%     \label{fig:c2mf}
% \end{figure}
% Figure 1: Credibility-Aware Fusion via Latent Context. Unlike static fusion methods, our framework extracts a joint neural embedding that serves as a context for a Conditional Probabilistic Circuit. The circuit dynamically evaluates the credibility of each source based on its position in the latent space (i.e., identifying when one modality is unreliable), ensuring that the final prediction is dominated by the most reliable information source for that specific instance. 

Recent deep learning based fusion architectures often rely on attention or gating mechanisms to reweight modalities. While powerful, these approaches typically produce opaque intermediate representations and do not provide an explicit probabilistic interpretation of modality credibility. In contrast, probabilistic circuit (PC) based fusion \cite{sidheekh2025credibility} has emerged as a promising way to combine information from multiple modalities effectively, while also enabling principled and probabilistic credibility measures via exact and tractable inference over predictive distributions. Such systems have also been shown to be more robust and reliable compared to neural counterparts, yielding well-calibrated classifiers, especially in noisy settings \cite{sidheekh2024robustness}. However, existing PC-based approaches model reliability as a static property, independently of any contextual variables, and are therefore unable to resolve context-specific conflicts.

In this work, we extend credibility-aware fusion to the \textbf{context-conditional} setting 
 and propose \textbf{C$^2$MF (Context-Specific Credibility Aware Multimodal Fusion)}, a hybrid framework that combines deep neural unimodal encoders with  Conditional Probabilistic Circuits (CPCs) \cite{shao2020conditional} as the fusion function. Our approach extracts a joint neural embedding from the input modalities and their corruptions, which serves as a latent context. This context is then used to parameterize the weights of the PC using a hyper-network, allowing the fusion logic to adapt dynamically for making reliable predictions, while preserving tractability for exact probabilistic inference and instance-specific credibility assessment. Figure \ref{alg:C2MF} illustrates the overall architecture. We also generalize the PC-based credibility
measure from~\cite{sidheekh2025credibility} to the context-conditional setting. We define \emph{Context-Specific Information Credibility} (CSIC) as the
KL divergence between the full multimodal posterior and the posterior with a modality marginalized out, both conditioned on the context $z$. This recovers the prior static credibility measure~\cite{sidheekh2025credibility}
as a special case when the context carries no additional information about modality reliability.
 To evaluate our C$^2$MF framework, we identify a lack of existing benchmarks for context-dependent conflict resolution and introduce the \textit{Conflict} benchmark. In this benchmark, we intentionally corrupt specific modalities based on class labels to induce structured cross-modal disagreement. We also propose the \emph{Reliable Modality Identification Score (RMIS)}, a metric that directly measures whether a model correctly identifies the more reliable modality under conflict.
% forcing the model to learn that "unreliability" is a function of the input's position in the latent space.

\noindent Thus, overall we make the following key contributions: \\
% \begin{enumerate}
%     \item We introduce C$^2$MF, a CPC-based fusion framework that dynamically conditions modality credibility on a learned latent context, enabling dynamic instance level reliability modeling while preserving exact probabilistic inference.
%     \item We generalize the probabilistic divergence based credibility measure definitions to this conditional setting.
%     \item We propose a structured corruption framework and introduce a new benchmark for evaluating fusion approaches on their ability to handle context-specific corruptions and cross-modal conflicts.
%     \item We propose RMIS, a new evaluation metric that directly quantifies a model's ability to identify the reliable modality under conflict.
%     \item We demonstrate through extensive experiments that $C^2MF$ substantially improves predictive performance and reliable modality identification compared to static PC baselines.
% \end{enumerate}
% (1) We introduce a probabilistic fusion framework that uses CPCs for dynamic, instance-level credibility modeling, (2) We introduce a new diagnostic dataset designed to evaluate multimodal models on their ability to handle context-specific source corruption and information conflict, (3) We demonstrate that our framework produces more accurate predictions compared to a PC-based static-reliability baseline, while its retaining tractability and explainability.
(1) We introduce C$^2$MF, a CPC-based fusion framework that dynamically conditions credibility inference on a learned latent context, enabling dynamic instance-level reliability modeling while preserving exact probabilistic semantics.
(2) We generalize the probabilistic divergence-based credibility measure definitions to this conditional setting.
(3) We propose a structured corruption framework and introduce a new benchmark for evaluating fusion approaches on their ability to handle context-specific corruptions and cross-modal conflicts.
(4) We propose RMIS, a new evaluation metric that directly quantifies a model's ability to identify the reliable modality under conflict.
(5) We demonstrate through extensive experiments that C$^2$MF substantially improves predictive performance and reliable modality identification compared to static baselines.

\section{Background and Related Work}

\subsection{Multimodal fusion and reliability}
Multimodal fusion\cite{hall1997sensorfusion, baltruvsaitis2018multimodal, liang2024foundations} aims to integrate information from $M$ distinct sources, $\mathbf{X} = \{X_1, \dots, X_M\}$ to improve the accuracy and robustness of a target prediction $Y \in \mathcal{Y}$. Formally, given a labelled dataset $\mathcal{D} = \{(x^{(i)}_1, \dots, x^{(i)}_M, y^{(i)})\}_{i=1}^{N}$ these methods aim to learn a function $f$ to model the conditional distribution over the target given multimodal data, i.e., $P(Y \mid x_1,\dots,x_M) = f(x_1,\dots,x_M).$ Fusion architectures are commonly classified as early (data-level), intermediate (feature-level), and late (decision-level).

\textbf{Early/Signal Fusion} \cite{snoek2005early, gadzicki2020early} integrates raw features or input data at the earliest stage, typically by concatenating $(x_1,\dots,x_M)$ into a single input vector and learning $P(Y \mid \mathbf{x}) = f(x_1 \oplus \dots \oplus x_M)$. While this allows the model to capture low-level correlations between sensors, it is highly sensitive to noise and makes reasoning about the contribution of individual sources difficult. 
% \textcolor{red}{Add citations}

\textbf{Late/Decision Fusion} processes each modality independently through  unimodal predictors  $f_m \in \mathcal{F}$, producing predictive distribution $p_m(y) = f_m(x_m)$, aggregating only the final decisions or logit scores \cite{xu2024reliable, zhang2023provable,han2021trusted} (e.g., via averaging or voting) \cite{ natarajan2005learning, tian2020uno}, yielding $P(Y \mid \mathbf{x}) = f(p_1, \ldots, p_M)$. The explicit separation of unimodal predictions makes late fusion well-suited for assessing per-modality credibility. It is also more robust to individual modality failures, but can fail to capture complex inter-modality interactions.

\textbf{Intermediate/Feature Fusion} \cite{ngiam2011multimodal} offers a middle ground by first extracting latent representations $h_m = E_m(x_m)$ from each modality and then fusing them through shared neural layers \cite{Joze2019MMTMMT,Zhang2019CPMNetsCP, perez2019mfas}. This is currently the dominant paradigm in deep learning, as it enables the learning of joint embeddings that represent high-level cross-modal concepts.
Recent advancements have moved toward \textbf{Hybrid Fusion}, which combines elements of multiple strategies—for instance, using intermediate fusion for feature extraction while maintaining separate decision branches for redundancy. These approaches construct an additional ``pseudo view" by combining features from two input modalities, and pass this as an auxiliary input for the fusion module \cite{han2022trusted, NEURIPS2021_371bce7d}. However, the primary challenge in any fusion scheme is the presence of modality conflict, where sensors provide contradictory evidence.

To resolve these conflicts, existing methods \cite{snidaro2016context} often incorporate meta-information regarding source reliability. 
This is typically achieved through credibility-aware frameworks that assign a weight $C_i$ to each modality based on its historical performance on a training set.
\cite{sidheekh2025credibility} formalizes \textbf{credibility} as the divergence between the full multimodal posterior $P(Y \mid p_1, \dots, p_M)$ and the posterior obtained by excluding modality $m$, and uses a Probabilistic Circuits (PC) to estimate this quantity tractably.
Such approaches assume that reliability is a stationary property. As a result, it cannot adapt to situations where the modality reliability varies with the input context.
% ; they treat a modality's credibility as a global constant rather than a context-dependent variable.
% Evaluating credibility requires a way to jointly reason about the predictive distributions

% Different types of fusion Late (decision), intermediate (feature), early (data). Hybrid

\subsection{Probabilistic Circuits as fusion functions}
Reliable multimodal fusion requires a combination function that can jointly model the predictive distributions from multiple sources and support efficient reasoning about their dependencies. Probabilistic Circuits
(PCs)~\cite{Choi20} are a class of generative models that are well suited to this role: they are expressive enough to capture complex dependencies between variables while supporting exact and efficient inference for a broad class of probabilistic queries \cite{sidheekh2024building}.

Formally, a PC represents a joint distribution $P(\mathbf{X})$ over a set of random variables $\mathbf{X}$ as a rooted Directed Acyclic Graph (DAG), whose leaf nodes represent simple univariate distributions and whose internal nodes are either sum nodes (representing weighted mixtures) or product nodes (representing factorizations over disjoint subsets of variables). Under mild structural constraints,
% \emph{smoothness} (all children of a sum node share the same scope) and \emph{decomposability} (all children of a product node have disjoint scopes) —
marginal and conditional distributions can be computed in time linear in the size of the circuit~\cite{SPNPoon2011}. This tractability is the key property that distinguishes PCs from general neural models.
In the context of late fusion, a PC can be used as the combination function by modeling the joint distribution over the unimodal predictive distributions $\{p_m\}_{m=1}^{M}$ and the target $Y$~\cite{sidheekh2025credibility}.
The fused prediction is then obtained by conditional inference: $P(Y \mid p_1, \dots, p_M) = P(Y, p_1, \dots, p_M) / P(p_1, \dots, p_M)$, which the PC computes exactly. Crucially, the same inference routines that yield the fused prediction also support the computation of credibility measures: the influence of modality $m$ can be quantified by comparing the full posterior $P(Y \mid p_1, \dots, p_M)$ against the posterior obtained by marginalizing $p_m$ out of the PC~\cite{sidheekh2025credibility}. This marginalization is handled natively by the circuit without requiring imputation or retraining, which also makes PCs naturally robust to missing modalities. PC-based fusion has been shown empirically to yield well-calibrated and reliable classifiers, particularly in noisy and incomplete data settings \cite{sidheekh2024robustness}.
In practice, PCs are instantiated using random tensorized structures \cite{peharz20a-rat-spn,peharz_20_einsum,sidheekh2023probabilistic} which are differentiable and support GPU-accelerated learning, making them straightforward to integrate into end-to-end deep learning pipelines.
% alongside neural unimodal encoders.

A key limitation of standard PCs in the fusion setting is that their parameters  are fixed after training, yielding a static fusion function. A growing line of work extends PCs to conditional distributions by allowing the circuit parameters to depend on an external input. Conditional SPNs
(CSPNs~\cite{shao2020conditional}) formalize this by predicting the sum-node weights from a conditioning variable $z$ via a neural gate function, yielding a tractable model of $P(\mathbf{Y} \mid \mathbf{X},z)$ that preserves exact inference. Related architectures include HyperSPNs~\cite{shih2021hyperspns}, which use hypernetworks to parameterize circuit weights, and $\chi$-SPNs~\cite{poonia2024chi} and Interventional SPNs~\cite{zevcevic2021interventional}, which extend this framework to causal settings. We refer to this family collectively as \emph{Conditional Probabilistic Circuits} (CPCs). In this work, we use a CPC as the fusion function, parameterized by context-conditioned weights, so that the fusion logic and the credibility estimates derived from it adapt dynamically to each input.

\section{Context-Specific Credibility-Aware Multimodal Fusion}
While multimodal fusion can yield more robust predictions, combining information from multiple modalities requires reasoning about their reliability. Source reliability \cite{costa2019urref} in real-world settings is often complex and context-dependent, yet existing approaches typically model reliability using static weights.
% and, as a result, struggle with scenarios involving context-specific corruption. 
We first formalize the problem of learning a fusion function
% modeling a predictive distribution that is 
robust to context-specific degradations in one or more modalities:
% . Formally, our problem can be expressed as:

\vspace{0.4em}
\noindent\fbox{%
    \parbox{.98\linewidth}{%
        \vspace{0.2cm}
        \begin{tabular}{lp{0.75\linewidth}}
            \textbf{Given:} & A dataset
            $\mathcal{D} = \{(\mathbf{x}^{(i)}, y^{(i)}), \textbf{k}^{(i)}\}_{i=1}^{N}$
            with inputs $\mathbf{x}^{(i)} = \{x^{(i)}_1, \dots, x^{(i)}_M\}$
            drawn from $M$ modalities, corruption masks $\mathbf{k}^{(i)} = \{k^{(i)}_1, \dots, k^{(i)}_M\}$  and a discrete target $Y \in \mathcal{Y}$. \\[4pt]
            \textbf{To Do:}  & Learn an accurate and explainable fusion function
            $\mathcal{P}$ that estimates $P(Y \mid x_1, \dots, x_M)$ while
            remaining robust to context-dependent corruptions.
        \end{tabular}
        \vspace{0.1cm}
    }%
}\\
% \noindent\fbox{%
%     \parbox{.95\linewidth}{%
%         \vspace{0.01cm}
%         \begin{tabular}{lp{0.75\linewidth}}
%             \textbf{Given:} & Input modalities $\mathbf{X} = \{X_1, \dots, x_M\}$, a discrete target $Y$, and a labeled dataset $\mathcal{D}$ \\
%             \textbf{To Do:}  & Learn an accurate yet explainable fusion function $F(\mathbf{X})$ that estimates $P(Y \mid x_1, \dots, x_M)$ while being robust to context-dependent source corruption.
%         \end{tabular}
%         \vspace{0.01cm}
%     }%
% }\\

\noindent We address this through a hybrid framework
$\text{C}^2\text{MF} = \langle \mathcal{E}, \mathcal{F}, \mathcal{M}, \mathcal{P} \rangle$, consisting of a set of unimodal encoders $\mathcal{E} = \{E_1, \dots, E_M\}$, a set of unimodal predictors $\mathcal{F} = \{f_1, \dots, f_M\}$, a joint context aggregator $\mathcal{M}$, and a context-specific late fusion module $\mathcal{P}$. We describe each component below in detail, followed by the learning procedure.
% \vspace{-0.05in}
\subsection{Unimodal Encoders and Predictors.}
\noindent Each modality $m$ is first processed by a dedicated encoder $E_m \in \mathcal{E}$
with parameters $\psi_m$, producing latent embeddings 
% $l_m = E_m(k_m)$ and 
$h_m = E_m(x_m)$.
This representation $h_m$ is passed to a unimodal predictor $f_m \in \mathcal{F}$ with parameters $\omega_m$, yielding the unimodal predictive distribution over the target: $   p_m = f_m(h_m)$. 
In addition to encoding the raw modality input, each encoder also produces a corruption embedding $l_m = E_m(k_m)$ that captures the pattern of corruption in each modality. In practice, the ground truth corruption mask $k_m$ need not be externally provided, and can be estimated from the data, making our framework applicable even when ground truth masks are unavailable.
The unimodal predictors are trained jointly with the rest of the framework via auxiliary supervision, ensuring that each $p_m$ carries discriminative information before fusion. 
% We address this problem through a hybrid fusion approach, factorizing $F$ into a set of unimodal encoders and predictors, a joint multimodal encoder, and a context-specific late-fusion function. Formally, $F =\langle \bm{E}, \bm{f}, M, P \rangle.$ Here, $\bm{E} = \{E_1,\dots,E_n\}$ is a set of unimodal encoders that process the corresponding modality in $\bm{X}.$ They produce a set of unimodal latent representations, $h_i = E(x_i), \forall i = 1,\dots,n.$ Each such representation is processed by a unimodal predictor $f_i \in \bm{f}$ to produce a distribution $p_i = f_i(h_i).$ Additionally, the latent representations are aggregated to form a multimodal embedding by the joint multimodal encoder $M.$ This multimodal embedding is combined with unimodal predictions by the context-specific late fusion module $P$ to output the final predictive distribution. We now describe each component in detail, and then present an algorithm to learn them from data.
% \vspace{-0.05in}
 \subsection{Joint Context Embedding.}
\noindent To satisfy (D1), i.e., inferring the operational context directly from the observed data, {\bf the framework must extract a representational summary that captures inter-modality correlations and the presence of situational corruptions}. Thus, we introduce a joint context aggregator $\mathcal{M}$, which operates on the collection of unimodal embeddings and produces:
$z = \mathcal{M}(h_1, \dots, h_M, l_1, \dots, l_M)$. Unlike the unimodal predictors, which learn modality-specific discriminative features, $\mathcal{M}$ is designed to capture how features in one modality may indicate the unreliability of another. For example, how high-frequency noise in an audio embedding implies that the audio stream should be down-weighted. The resulting embedding $z$ serves as the conditioning input for the CPC fusion stage. 
In practice, we implement $\mathcal{M}$ as a lightweight MLP applied to the concatenation of unimodal embeddings.
% , keeping the context aggregation step computationally negligible relative to the unimodal encoders.
% \subsection{Joint multimodal embedding}
% To satisfy the requirement of sensing situational corruption (D1), the framework must extract a representational summary of the environment that goes beyond individual modality predictions. We introduce a joint context aggregator, which operates on the collection of latent representations $h_1,\dots,h_n.$

% Unlike the unimodal predictors, which learn modality-specific discriminative features, the joint context-aggregator is designed to identify inter-modality correlations and environmental noise. As a result, it must learn to capture how the presence of certain features in one modality (e.g., high-frequency audio noise) might imply the unreliability of another. Formally, we compute the joint context embedding $z$ as 
% $$z = g_\phi(h_1,\dots,h_n).$$

% This joint embedding serves as the necessary precondition for the dynamic reweighting of the fusion logic in the subsequent stage.

\subsection{Fusion via Conditional Probabilistic Circuits.}
To enable instance-specific reliability modeling (D2), we use the joint context $z$ to parameterize the fusion logic. We map this context to the internal parameters of a Conditional Probabilistic Circuit (CPC) through a hyper-network, $g_\phi.$ Specifically, the hyper-network computes the sum-node weights $\Theta$ for the CPC as
$\Theta = g_\phi(z)$.
In this architecture, the CPC acts as a {\bf differentiable and tractable fusion operator} that models the joint distribution over the unimodal predictions $\{p_m\}_{m=1}^{M}$ and the target variable $Y$. While its product nodes encode independence assumptions between unimodal predictions, its sum nodes represent mixtures over them. By conditioning the sum weights $\Theta$ on the joint context $z,$ the hyper-network can ``gate" the circuit's branches in real-time. For example, if the joint context $z$ indicates significant corruption in one modality, $g_\phi$ generates weights that attenuate its influence and prioritize the other modalities. 
Crucially, this mechanism satisfies D3 (Tractable and Explainable Reasoning). Unlike neural intermediate fusion functions, which are black-box models, the resulting Conditional Probabilistic Circuit (CPC) represents a valid probability distribution. This allows the framework to perform exact inference of the global posterior $P(Y \mid X,z)$ by propagating the unimodal predictions $\{p_1,\dots,p_n\}$ through the circuit in a single bottom-up pass.

\subsection{Context-specific Information credibility}
% While the CPC can dynamically resolve conflicts, 
A critical requirement for high-stakes applications is the ability to audit why a specific decision was reached (D3). To provide this transparency, we introduce \textit{Context-Specific Information Credibility (CSIC)}. {\bf This metric quantifies the influence of each modality's information on the final prediction under the current environmental context}.

Formally, we define the CSIC for information obtained from modality $i$ as the Kullback-Leibler (KL) divergence between the full multimodal posterior distribution over the target and the posterior where modality $i$ has been marginalized out, both conditioned on the joint context $z$:
% We extend the framework of credibility-aware fusion\cite{sidheekh2025credibility} to the conditional setting. We define the Context-Specific Information Credibility (CSIC) for modality i as the KL divergence between the full posterior and the posterior where modality i is marginalized out, conditioned on context z:
$$ CSIC_i({\bf x, z}) = D_{KL}(P(Y | {\bf x, z})\ ||\ P(Y | {\bf x}\setminus x_i, {\bf z})),$$
% $$CSIC_i = \frac{CSIC_i}{\sum_i^M{CSIC_i}}$$
which can be computed exactly using the CPC's tractable inference
routines.  The relative
CSIC score normalizes these values across modalities:
$$
    \overline{CSIC}_i(\mathbf{x}, \mathbf{z})
    = \frac{CSIC_i(\mathbf{x}, \mathbf{z})}
           {\sum_{j=1}^{M} CSIC_j(\mathbf{x}, \mathbf{z})},
$$
which satisfies $\overline{CSIC}_m \in [0, 1]$ and
$\sum_m \overline{\mathrm{CSIC}}_m = 1$, giving a normalized
instance-level credibility distribution over modalities. $\overline{CSIC}$ generalizes
the static credibility measure of~\cite{sidheekh2025credibility}
% Note that CSIC generalizes the Credibility score\cite{sidheekh2025credibility}, 
which is recovered as a special case when the joint context $z$ provides no additional information about information credibility, i.e., a domain without context-specific corruption.

\subsection{Learning}
\noindent To ensure that the joint context aggregator and the CPC-based fusion logic work in harmony, we optimize the entire framework end-to-end. The learning objective must encourage the unimodal encoders to extract discriminative features while simultaneously training the hyper-network to recognize which features signify unreliability.
We thus  define a joint training objective that combines the multimodal fusion loss with auxiliary unimodal supervision: $\underset{\Omega}{\arg \min} \ \mathcal{L}_{\mathrm{total}}
= \mathcal{L}_{f} + \mu\, \mathcal{L}_{u},$
% We  define a joint loss function $\mathcal{L}_\text{total}$ that combines a multimodal fusion loss with auxiliary unimodal supervision:
% $$\underset{\Omega}{\arg \min}\ \frac{1}{|\mathcal{D}|}\sum_{(x, y) \in \mathcal{D}} \ell (y, P(Y \mid X = x))$$
% where $\ell$ denotes the cross-entropy loss,  and $\Omega = \{\psi, \phi, M, g, \theta\}$ is the set of all learnable parameters in the architecture.
% During training, the gradient $\nabla \mathcal{L}_\text{total}$ is backpropagated through the CPC to the hyper-network $g_\phi$ and the context aggregator $\mathcal{M}$. This allows the context aggregator to learn a latent space where situational corruptions are easily identifiable. As a result, the hyper-network learns to map these "noisy" regions of the latent space to sum-node weights that down-weight the unreliable modality. The complete optimization procedure is detailed in Algorithm 1.
where
\begin{align}
    \mathcal{L}_{f} &= \frac{1}{|\mathcal{D}|}
        \sum_{(\mathbf{x}, y) \in \mathcal{D}}
        \ell\!\left(y,\, P(Y \mid X = x)\right), \\
    \mathcal{L}_{u} &= \frac{1}{|\mathcal{D}|}
        \sum_{(\mathbf{x}, y) \in \mathcal{D}}
        \sum_{m=1}^{M} \ell\!\left(y,\, p_m\right),
\end{align}
$\ell$ denotes the cross-entropy loss, and $\mu$ is a weighting
hyperparameter. The set of all learnable parameters is
$\Omega = \{\psi, \phi, \omega_{\mathcal{F}}\}$, where
$\psi = \{\psi_m\}_{m=1}^{M}$ are the encoder parameters,
$\phi$ are the hyper-network parameters, and
$\omega_{\mathcal{F}} = \{\omega_m\}_{m=1}^{M}$ are the unimodal
predictor parameters. 
The gradient $\nabla \mathcal{L}_\text{total}$ is backpropagated through the CPC to the hyper-network $g_\phi$, the context aggregator $\mathcal{M}$, and the encoders $\mathcal{E}$. This allows the context aggregator to learn a latent space where situational corruptions are easily identifiable. As a result, the hyper-network learns to map these ``noisy" regions of the latent space to sum-node weights that down-weight the unreliable modality. The complete optimization procedure is detailed in Alg. 1.
% Gradients of $\mathcal{L}_{\mathrm{total}}$ are
% backpropagated through the CPC to $g_\phi$ and the encoders $\mathcal{E}$,
% encouraging the encoder representations to carry the contextual signal
% needed for $g_\phi$ to identify and down-weight unreliable modalities.
% The complete procedure is detailed in Algorithm~\ref{alg:C2MF}.

\begin{algorithm}[!t]
\caption{$\text{C}^2\text{MF}$ Training}
\label{alg:C2MF}
\SetInd{0.4em}{0.4em}
\small
\SetKwInOut{KwIn}{In}
\SetKwInOut{KwOut}{Out}
\KwIn{Dataset $\mathcal{D}$; Encoders $\mathcal{E} = \{E_m\}_{m=1}^{M}$
      with parameters $\psi$; Predictors $\mathcal{F} = \{f_m\}_{m=1}^{M}$
      with parameters $\omega_{\mathcal{F}}$; Context aggregator $\mathcal{M}$;
      Hyper-network $g_\phi$; CPC $\mathcal{P}_\Theta$}
\KwOut{Parameters $\Omega = \{\psi, \phi, \omega_{\mathcal{F}}\}$}

\While{not converged}{
    Sample batch
    $\mathcal{B} = \{(\mathbf{x}^{(i)}, y^{(i)})\}_{i=1}^{B} \sim \mathcal{D}$\;

    \For{$m = 1$ \KwTo $M$}{
        $h_{m} \leftarrow E_m(x_{m})$ \tcp*{Unimodal embeddings}
        $l_{m} \leftarrow E_m(k_{m})$ \tcp*{Unimodal corruption embeddings}
        $p_{m} \leftarrow f_m(h_{m})$ \tcp*{Unimodal predictions}
    }

    $z \leftarrow \mathcal{M}(h_1, \dots, h_M, l_1, \dots, l_M)$ \tcp*{Joint context}
    $\Theta \leftarrow g_\phi(z)$ \tcp*{Context-conditioned weights}

    $\hat{y} \leftarrow \mathcal{P}(Y|p_1, \dots, p_M;\, \Theta)$ \tcp*{ Fused output}

    $\mathcal{L}_{f} \leftarrow \frac{1}{B}
        \sum_{i \in \mathcal{B}} \ell(\hat{y}^{(i)}, y^{(i)})$\;
    $\mathcal{L}_{u} \leftarrow \frac{1}{B}
        \sum_{i \in \mathcal{B}} \sum_{m=1}^{M} \ell(p_m^{(i)}, y^{(i)})$\;
    $\mathcal{L}_{\mathrm{total}} \leftarrow
        \mathcal{L}_{f} + \mu\, \mathcal{L}_{u}$\;

    $\Omega \leftarrow \Omega - \eta\, \nabla_\Omega
        \mathcal{L}_{\mathrm{total}}$ \tcp*{Update all parameters}
}
% \vspace{-0.2in}
\end{algorithm}

\section{Experimental Setup}

% We aim to explicitly answer the following research questions:
\noindent We design our evaluation around two key research questions:
\begin{itemize}
    \item[{\bf (Q1)}] Does context-specific credibility-aware fusion yield more accurate predictions than static credibility-aware fusion? %(Case 1: CWM, Case 2: Full-PC)
    \item[{\bf (Q2)}] In the presence of cross-modal conflicts, can C$^2$MF correctly distinguish between a unimodal predictor being confident and the modality being reliable? %[New metrics]
    % \item[{\bf (Q3)}] How does the complexity of the joint embedding encoder affect performance?
    % \item joint training
    % \item calibration
\end{itemize}

\subsection{The Conflict Benchmark}
To systematically evaluate the model's ability to identify the reliable modality under cross-modal disagreement, we first introduce a controlled conflict generation framework. This framework explicitly injects conflicts into multimodal datasets by corrupting one modality while keeping the other intact. 

Formally, for each modality $m$, we define a subset of classes $c_m$ whose samples are designated for corruption. %These subsets are constructed to be strictly non-overlapping. % to ensure that for a particular class, we shift the reliability of the model away from the corrupted modality. 
For each class $c \in c_m$, we select a fraction $\lambda \in [0, 1]$ of its example, and corrupt each example by replacing part (or the entirety) of the modality-specific information with that of another example sampled from a predefined target class $c' \notin c_m$. This procedure introduces controlled inconsistency between the modalities, thereby deliberately decreasing the credibility of the information belonging to the corrupted modality, rendering it unreliable. 
Although the framework is generally applicable to any multimodal dataset, we currently provide support for the following datasets: AV-MNIST and NYUD.
\subsubsection{Conflict-AV-MNIST}
AV-MNIST (Audio-Visual MNIST) \cite{vielzeuf2018centralnet} is a multimodal benchmark constructed from handwritten digit images (0–9) from MNIST paired with corresponding audio recordings from the Free Spoken Digit Dataset (FSDD) \cite{zohar_jackson_2018_1342401}. Each audio clip is converted into a Mel-Frequency Cepstral Coefficient (MFCC) representation of size $112 \times 112$. This dataset contains 55,000 training, 5,000 validation, and 10,000 test examples.
We construct Conflict-AV-MNIST by introducing noise into class sets $c_\text{image}  = \{5, 8, 1\}$ and $c_\text{audio} = \{6, 0, 7\}$ for image and audio modalities respectively. We set a fixed conflict ratio of $\lambda = 0.7$ on the train data. That is, $70\%$ of the images from classes $c_\text{image} = \{5, 8, 1\}$ are replaced with that of train examples from classes $c' \notin c_\text{image}$. Similarly, for each class $c \in c_\text{audio}$, $70\%$ of its audio samples are replaced with audio drawn from classes $c' \notin c_\text{audio}$.

\subsubsection{Conflict-NYUD}
NYU Depth (NYUD) \cite{silberman2012indoor} is an RGB-D indoor scene classification benchmark consisting of paired RGB and depth images. Following \cite{zhang2023provable}, we use a subset of this dataset restricted to 10 classes, comprising 1,863 RGB-depth image pairs split into 795 train, 414 validation, and 654 test examples.
We construct the conflict version of NYUD by setting $\lambda_\text{train}$ to $0.7$, $c_\text{RGB} = \{0, 5, 7\}$ and $c_\text{Depth} = \{2, 6, 9\}$.

\subsection{Baselines}
We consider the naive context-agnostic late fusion methods \cite{sidheekh2025credibility} as baselines for our experimental evaluation. We describe each of them very briefly here.
\subsubsection{Direct-PC}
Direct-PC (DPC)\cite{sidheekh2025credibility} is the context-agnostic fusion method that uses a Probabilistic Circuit as a fusion function. It takes the unimodal predictions as input and computes the joint $P(Y,  p_1..p_m)$. The multimodal predictive distribution is computed via conditional inference as 
\begin{equation*}
    P(Y\mid p_1..p_m) = \frac{P(Y,p_1..p_m)}{P(p_1..p_m)}
\end{equation*}
\subsubsection{Credibility Weighted Mean}
Credibility Weighted Mean (CWM) derives the per-modality credibility scores from the joint distribution learned by the static PC as 
$$C_i = D_{KL}(P(Y | {\bf x})\ ||\ P(Y | {\bf x}\setminus x_i),$$ and the relative credibility as
$$ \overline{C_i} = \frac{C_i}{\sum_i^M C_i}$$
The multimodal predictive distribution is computed as a credibility-weighted average of the unimodal predictive distributions, i.e. 
$$P(Y\mid p_1..p_m) = \sum_{i}^{M} \overline{C_i}p_i$$
    
Correspondingly, we define our \textbf{context-specific variations} of the above baselines as \textbf{C$^2$DPC} and \textbf{C$^2$WM}, where C$^2$DPC extends DPC and C$^2$WM generalizes CWM by replacing the static PC with a CPC parameterized by $\Theta=g_\phi(h_1,\ldots,h_M)$, enabling credibility estimates to adapt dynamically.

\subsection{Metrics}
We measure each model's performance using the classification accuracy, precision, recall, and F1 score. To measure the ability of the models to correctly identify the reliable modality under conflict, we introduce the Reliable Modality Identification Score (RMIS) as follows:
\begin{equation*}
    RMIS^i = \begin{cases}
            0, &\quad {CSIC}^i_{m_\text{corrupt}} > {CSIC}^i_{m_\text{clean}} \\
            1, &\quad \text{otherwise} \\
            \end{cases}
\end{equation*}
where $m_\text{corrupt}$ and $m_\text{clean}$ denote the corrupted and clean
modalities for instance $i$ respectively. A model achieves $\mathrm{RMIS} = 1$
on an instance, if and only if it assigns higher credibility to the clean
modality than to the corrupted one.

\subsection{Results}

\begin{table*}[!ht]
\centering
\small
\setlength{\tabcolsep}{4pt}
\renewcommand{\arraystretch}{1.15}

\subcaptionbox{Accuracy\label{tab:accuracy}}{%
\begin{tabular}{ll | cccc | cccc}
\hline
\multirow{2}{*}{\textbf{Training}} & \multirow{2}{*}{\textbf{Method}}
  & \multicolumn{4}{c|}{\textbf{Conflict-AV-MNIST}}
  & \multicolumn{4}{c}{\textbf{Conflict-NYUD}} \\ \cline{3-10}
 & & \textbf{0\%} & \textbf{50\%} & \textbf{75\%} & \textbf{100\%}
   & \textbf{0\%} & \textbf{50\%} & \textbf{75\%} & \textbf{100\%} \\ \hline
\multirow{4}{*}{Decoupled}
 & \cellcolor{cwmgroup}CWM
   & \gradientcell{98.91}{70}{100}{cwmgroupdark}{}
   & \gradientcell{84.73}{70}{100}{cwmgroupdark}{}
   & \gradientcell{77.61}{70}{100}{cwmgroupdark}{}
   & \gradientcell{70.51}{70}{100}{cwmgroupdark}{}
   & \gradientcell{54.13}{45}{70}{cwmgroupdark}{}
   & \gradientcell{53.52}{45}{70}{cwmgroupdark}{}
   & \gradientcell{53.01}{45}{70}{cwmgroupdark}{}
   & \gradientcell{52.65}{45}{70}{cwmgroupdark}{} \\
 & \cellcolor{cwmgroup}\textbf{C$^2$WM}
   & \gradientcell{97.65}{70}{100}{cwmgroupdark}{-1.26}
   & \gradientcell{97.38}{70}{100}{cwmgroupdark}{12.65}
   & \gradientcell{97.25}{70}{100}{cwmgroupdark}{19.64}
   & \gradientcell{97.04}{70}{100}{cwmgroupdark}{26.53}
   & \gradientcell{53.77}{45}{70}{cwmgroupdark}{-0.36}
   & \gradientcell{53.16}{45}{70}{cwmgroupdark}{-0.36}
   & \gradientcell{53.31}{45}{70}{cwmgroupdark}{0.30}
   & \gradientcell{53.31}{45}{70}{cwmgroupdark}{0.66} \\ \cline{2-10}
 & \cellcolor{dpcgroup}DPC
   & \gradientcell{98.92}{70}{100}{dpcgroupdark}{}
   & \gradientcell{84.40}{70}{100}{dpcgroupdark}{}
   & \gradientcell{77.24}{70}{100}{dpcgroupdark}{}
   & \gradientcell{70.09}{70}{100}{dpcgroupdark}{}
   & \gradientcell{49.90}{45}{70}{dpcgroupdark}{}
   & \gradientcell{50.00}{45}{70}{dpcgroupdark}{}
   & \gradientcell{49.34}{45}{70}{dpcgroupdark}{}
   & \gradientcell{49.13}{45}{70}{dpcgroupdark}{} \\
 & \cellcolor{dpcgroup}\textbf{C$^2$DPC}
   & \gradientcell{99.14}{70}{100}{dpcgroupdark}{0.22}
   & \gradientcell{99.04}{70}{100}{dpcgroupdark}{14.64}
   & \gradientcell{98.99}{70}{100}{dpcgroupdark}{21.75}
   & \gradientcell{98.95}{70}{100}{dpcgroupdark}{28.86}
   & \gradientcell{55.86}{45}{70}{dpcgroupdark}{5.96}
   & \gradientcell{59.38}{45}{70}{dpcgroupdark}{9.38}
   & \gradientcell{61.52}{45}{70}{dpcgroupdark}{12.18}
   & \gradientcell{64.22}{45}{70}{dpcgroupdark}{15.09} \\ \hline
\multirow{4}{*}{Joint}
 & \cellcolor{cwmgroup}CWM
   & \gradientcell{98.74}{70}{100}{cwmgroupdark}{}
   & \gradientcell{84.35}{70}{100}{cwmgroupdark}{}
   & \gradientcell{77.29}{70}{100}{cwmgroupdark}{}
   & \gradientcell{70.09}{70}{100}{cwmgroupdark}{}
   & \gradientcell{55.15}{45}{65}{cwmgroupdark}{}
   & \gradientcell{54.18}{45}{65}{cwmgroupdark}{}
   & \gradientcell{54.33}{45}{65}{cwmgroupdark}{}
   & \gradientcell{53.98}{45}{65}{cwmgroupdark}{} \\
 & \cellcolor{cwmgroup}\textbf{C$^2$WM}
   & \gradientcell{98.64}{70}{100}{cwmgroupdark}{-0.10}
   & \gradientcell{97.91}{70}{100}{cwmgroupdark}{13.56}
   & \gradientcell{97.70}{70}{100}{cwmgroupdark}{20.41}
   & \gradientcell{97.42}{70}{100}{cwmgroupdark}{27.33}
   & \gradientcell{58.00}{45}{65}{cwmgroupdark}{2.85}
   & \gradientcell{57.34}{45}{65}{cwmgroupdark}{3.16}
   & \gradientcell{56.83}{45}{65}{cwmgroupdark}{2.50}
   & \gradientcell{56.07}{45}{65}{cwmgroupdark}{2.09} \\ \cline{2-10}
 & \cellcolor{dpcgroup}DPC
   & \gradientcell{99.00}{70}{100}{dpcgroupdark}{}
   & \gradientcell{84.56}{70}{100}{dpcgroupdark}{}
   & \gradientcell{77.44}{70}{100}{dpcgroupdark}{}
   & \gradientcell{70.05}{70}{100}{dpcgroupdark}{}
   & \gradientcell{50.15}{45}{65}{dpcgroupdark}{}
   & \gradientcell{52.50}{45}{65}{dpcgroupdark}{}
   & \gradientcell{52.70}{45}{65}{dpcgroupdark}{}
   & \gradientcell{53.67}{45}{65}{dpcgroupdark}{} \\
 & \cellcolor{dpcgroup}\textbf{C$^2$DPC}
   & \gradientcell{99.24}{70}{100}{dpcgroupdark}{0.24}
   & \gradientcell{99.46}{70}{100}{dpcgroupdark}{14.90}
   & \gradientcell{99.53}{70}{100}{dpcgroupdark}{22.09}
   & \gradientcell{99.61}{70}{100}{dpcgroupdark}{29.56}
   & \gradientcell{59.28}{45}{65}{dpcgroupdark}{9.13}
   & \gradientcell{60.14}{45}{65}{dpcgroupdark}{7.64}
   & \gradientcell{59.89}{45}{65}{dpcgroupdark}{7.19}
   & \gradientcell{61.01}{45}{65}{dpcgroupdark}{7.34} \\ \hline
\end{tabular}}

\vspace{0.8em}

\subcaptionbox{F1-Score\label{tab:f1}}{%
\begin{tabular}{ll | cccc | cccc}
\hline
\multirow{2}{*}{\textbf{Training}} & \multirow{2}{*}{\textbf{Method}}
  & \multicolumn{4}{c|}{\textbf{Conflict-AV-MNIST}}
  & \multicolumn{4}{c}{\textbf{Conflict-NYUD}} \\ \cline{3-10}
 & & \textbf{0\%} & \textbf{50\%} & \textbf{75\%} & \textbf{100\%}
   & \textbf{0\%} & \textbf{50\%} & \textbf{75\%} & \textbf{100\%} \\ \hline
\multirow{4}{*}{Decoupled}
 & \cellcolor{cwmgroup}CWM
   & \gradientcell{98.89}{60}{100}{cwmgroupdark}{}
   & \gradientcell{83.60}{60}{100}{cwmgroupdark}{}
   & \gradientcell{74.00}{60}{100}{cwmgroupdark}{}
   & \gradientcell{61.43}{60}{100}{cwmgroupdark}{}
   & \gradientcell{40.56}{35}{65}{cwmgroupdark}{}
   & \gradientcell{39.93}{35}{65}{cwmgroupdark}{}
   & \gradientcell{39.54}{35}{65}{cwmgroupdark}{}
   & \gradientcell{39.22}{35}{65}{cwmgroupdark}{} \\
 & \cellcolor{cwmgroup}\textbf{C$^2$WM}
   & \gradientcell{97.63}{60}{100}{cwmgroupdark}{-1.26}
   & \gradientcell{97.36}{60}{100}{cwmgroupdark}{13.76}
   & \gradientcell{97.21}{60}{100}{cwmgroupdark}{23.21}
   & \gradientcell{97.00}{60}{100}{cwmgroupdark}{35.57}
   & \gradientcell{41.76}{35}{65}{cwmgroupdark}{1.20}
   & \gradientcell{40.09}{35}{65}{cwmgroupdark}{0.16}
   & \gradientcell{40.80}{35}{65}{cwmgroupdark}{1.26}
   & \gradientcell{40.29}{35}{65}{cwmgroupdark}{1.07} \\ \cline{2-10}
 & \cellcolor{dpcgroup}DPC
   & \gradientcell{98.88}{60}{100}{dpcgroupdark}{}
   & \gradientcell{83.53}{60}{100}{dpcgroupdark}{}
   & \gradientcell{74.72}{60}{100}{dpcgroupdark}{}
   & \gradientcell{63.83}{60}{100}{dpcgroupdark}{}
   & \gradientcell{38.33}{35}{65}{dpcgroupdark}{}
   & \gradientcell{38.14}{35}{65}{dpcgroupdark}{}
   & \gradientcell{38.27}{35}{65}{dpcgroupdark}{}
   & \gradientcell{37.28}{35}{65}{dpcgroupdark}{} \\
 & \cellcolor{dpcgroup}\textbf{C$^2$DPC}
   & \gradientcell{99.10}{60}{100}{dpcgroupdark}{0.22}
   & \gradientcell{99.01}{60}{100}{dpcgroupdark}{15.48}
   & \gradientcell{98.97}{60}{100}{dpcgroupdark}{24.25}
   & \gradientcell{98.93}{60}{100}{dpcgroupdark}{35.10}
   & \gradientcell{50.40}{35}{65}{dpcgroupdark}{12.07}
   & \gradientcell{53.14}{35}{65}{dpcgroupdark}{15.00}
   & \gradientcell{55.46}{35}{65}{dpcgroupdark}{17.19}
   & \gradientcell{58.53}{35}{65}{dpcgroupdark}{21.25} \\ \hline
\multirow{4}{*}{Joint}
 & \cellcolor{cwmgroup}CWM
   & \gradientcell{98.72}{60}{100}{cwmgroupdark}{}
   & \gradientcell{83.37}{60}{100}{cwmgroupdark}{}
   & \gradientcell{74.29}{60}{100}{cwmgroupdark}{}
   & \gradientcell{62.65}{60}{100}{cwmgroupdark}{}
   & \gradientcell{42.59}{35}{60}{cwmgroupdark}{}
   & \gradientcell{40.81}{35}{60}{cwmgroupdark}{}
   & \gradientcell{40.09}{35}{60}{cwmgroupdark}{}
   & \gradientcell{39.76}{35}{60}{cwmgroupdark}{} \\
 & \cellcolor{cwmgroup}\textbf{C$^2$WM}
   & \gradientcell{98.60}{60}{100}{cwmgroupdark}{-0.12}
   & \gradientcell{97.85}{60}{100}{cwmgroupdark}{14.48}
   & \gradientcell{97.62}{60}{100}{cwmgroupdark}{23.33}
   & \gradientcell{97.31}{60}{100}{cwmgroupdark}{34.66}
   & \gradientcell{44.43}{35}{60}{cwmgroupdark}{1.84}
   & \gradientcell{43.24}{35}{60}{cwmgroupdark}{2.43}
   & \gradientcell{41.88}{35}{60}{cwmgroupdark}{1.79}
   & \gradientcell{41.86}{35}{60}{cwmgroupdark}{2.10} \\ \cline{2-10}
 & \cellcolor{dpcgroup}DPC
   & \gradientcell{98.98}{60}{100}{dpcgroupdark}{}
   & \gradientcell{83.59}{60}{100}{dpcgroupdark}{}
   & \gradientcell{74.48}{60}{100}{dpcgroupdark}{}
   & \gradientcell{62.39}{60}{100}{dpcgroupdark}{}
   & \gradientcell{37.19}{35}{60}{dpcgroupdark}{}
   & \gradientcell{39.10}{35}{60}{dpcgroupdark}{}
   & \gradientcell{39.61}{35}{60}{dpcgroupdark}{}
   & \gradientcell{40.06}{35}{60}{dpcgroupdark}{} \\
 & \cellcolor{dpcgroup}\textbf{C$^2$DPC}
   & \gradientcell{99.22}{60}{100}{dpcgroupdark}{0.24}
   & \gradientcell{99.45}{60}{100}{dpcgroupdark}{15.86}
   & \gradientcell{99.52}{60}{100}{dpcgroupdark}{25.04}
   & \gradientcell{99.60}{60}{100}{dpcgroupdark}{37.21}
   & \gradientcell{53.50}{35}{60}{dpcgroupdark}{16.31}
   & \gradientcell{53.27}{35}{60}{dpcgroupdark}{14.17}
   & \gradientcell{53.11}{35}{60}{dpcgroupdark}{13.50}
   & \gradientcell{52.70}{35}{60}{dpcgroupdark}{12.64} \\ \hline
\end{tabular}}

\vspace{0.8em}

\subcaptionbox{Precision\label{tab:precision}}{%
\begin{tabular}{ll | cccc | cccc}
\hline
\multirow{2}{*}{\textbf{Training}} & \multirow{2}{*}{\textbf{Method}}
  & \multicolumn{4}{c|}{\textbf{Conflict-AV-MNIST}}
  & \multicolumn{4}{c}{\textbf{Conflict-NYUD}} \\ \cline{3-10}
 & & \textbf{0\%} & \textbf{50\%} & \textbf{75\%} & \textbf{100\%}
   & \textbf{0\%} & \textbf{50\%} & \textbf{75\%} & \textbf{100\%} \\ \hline
\multirow{4}{*}{Decoupled}
 & \cellcolor{cwmgroup}CWM
   & \gradientcell{98.94}{65}{100}{cwmgroupdark}{}
   & \gradientcell{88.77}{65}{100}{cwmgroupdark}{}
   & \gradientcell{84.31}{65}{100}{cwmgroupdark}{}
   & \gradientcell{68.34}{65}{100}{cwmgroupdark}{}
   & \gradientcell{40.49}{35}{65}{cwmgroupdark}{}
   & \gradientcell{39.52}{35}{65}{cwmgroupdark}{}
   & \gradientcell{39.24}{35}{65}{cwmgroupdark}{}
   & \gradientcell{39.10}{35}{65}{cwmgroupdark}{} \\
 & \cellcolor{cwmgroup}\textbf{C$^2$WM}
   & \gradientcell{97.77}{65}{100}{cwmgroupdark}{-1.17}
   & \gradientcell{97.49}{65}{100}{cwmgroupdark}{8.72}
   & \gradientcell{97.37}{65}{100}{cwmgroupdark}{13.06}
   & \gradientcell{97.20}{65}{100}{cwmgroupdark}{28.86}
   & \gradientcell{46.04}{35}{65}{cwmgroupdark}{5.55}
   & \gradientcell{40.77}{35}{65}{cwmgroupdark}{1.25}
   & \gradientcell{42.41}{35}{65}{cwmgroupdark}{3.17}
   & \gradientcell{40.93}{35}{65}{cwmgroupdark}{1.83} \\ \cline{2-10}
 & \cellcolor{dpcgroup}DPC
   & \gradientcell{98.93}{65}{100}{dpcgroupdark}{}
   & \gradientcell{86.90}{65}{100}{dpcgroupdark}{}
   & \gradientcell{81.26}{65}{100}{dpcgroupdark}{}
   & \gradientcell{67.76}{65}{100}{dpcgroupdark}{}
   & \gradientcell{44.04}{35}{65}{dpcgroupdark}{}
   & \gradientcell{44.23}{35}{65}{dpcgroupdark}{}
   & \gradientcell{44.88}{35}{65}{dpcgroupdark}{}
   & \gradientcell{42.63}{35}{65}{dpcgroupdark}{} \\
 & \cellcolor{dpcgroup}\textbf{C$^2$DPC}
   & \gradientcell{99.14}{65}{100}{dpcgroupdark}{0.21}
   & \gradientcell{99.05}{65}{100}{dpcgroupdark}{12.15}
   & \gradientcell{99.03}{65}{100}{dpcgroupdark}{17.77}
   & \gradientcell{99.01}{65}{100}{dpcgroupdark}{31.25}
   & \gradientcell{53.23}{35}{65}{dpcgroupdark}{9.19}
   & \gradientcell{55.31}{35}{65}{dpcgroupdark}{11.08}
   & \gradientcell{58.04}{35}{65}{dpcgroupdark}{13.16}
   & \gradientcell{61.03}{35}{65}{dpcgroupdark}{18.40} \\ \hline
\multirow{4}{*}{Joint}
 & \cellcolor{cwmgroup}CWM
   & \gradientcell{98.79}{65}{100}{cwmgroupdark}{}
   & \gradientcell{87.88}{65}{100}{cwmgroupdark}{}
   & \gradientcell{83.01}{65}{100}{cwmgroupdark}{}
   & \gradientcell{66.46}{65}{100}{cwmgroupdark}{}
   & \gradientcell{46.79}{35}{60}{cwmgroupdark}{}
   & \gradientcell{41.42}{35}{60}{cwmgroupdark}{}
   & \gradientcell{39.15}{35}{60}{cwmgroupdark}{}
   & \gradientcell{38.96}{35}{60}{cwmgroupdark}{} \\
 & \cellcolor{cwmgroup}\textbf{C$^2$WM}
   & \gradientcell{98.64}{65}{100}{cwmgroupdark}{-0.15}
   & \gradientcell{97.95}{65}{100}{cwmgroupdark}{10.07}
   & \gradientcell{97.80}{65}{100}{cwmgroupdark}{14.79}
   & \gradientcell{97.62}{65}{100}{cwmgroupdark}{31.16}
   & \gradientcell{46.95}{35}{60}{cwmgroupdark}{0.16}
   & \gradientcell{43.93}{35}{60}{cwmgroupdark}{2.51}
   & \gradientcell{41.75}{35}{60}{cwmgroupdark}{2.60}
   & \gradientcell{43.14}{35}{60}{cwmgroupdark}{4.18} \\ \cline{2-10}
 & \cellcolor{dpcgroup}DPC
   & \gradientcell{99.00}{65}{100}{dpcgroupdark}{}
   & \gradientcell{87.73}{65}{100}{dpcgroupdark}{}
   & \gradientcell{82.64}{65}{100}{dpcgroupdark}{}
   & \gradientcell{66.52}{65}{100}{dpcgroupdark}{}
   & \gradientcell{41.96}{35}{60}{dpcgroupdark}{}
   & \gradientcell{46.96}{35}{60}{dpcgroupdark}{}
   & \gradientcell{45.94}{35}{60}{dpcgroupdark}{}
   & \gradientcell{46.77}{35}{60}{dpcgroupdark}{} \\
 & \cellcolor{dpcgroup}\textbf{C$^2$DPC}
   & \gradientcell{99.25}{65}{100}{dpcgroupdark}{0.25}
   & \gradientcell{99.46}{65}{100}{dpcgroupdark}{11.73}
   & \gradientcell{99.53}{65}{100}{dpcgroupdark}{16.89}
   & \gradientcell{99.61}{65}{100}{dpcgroupdark}{33.09}
   & \gradientcell{55.59}{35}{60}{dpcgroupdark}{13.63}
   & \gradientcell{55.90}{35}{60}{dpcgroupdark}{8.94}
   & \gradientcell{54.53}{35}{60}{dpcgroupdark}{8.59}
   & \gradientcell{54.94}{35}{60}{dpcgroupdark}{8.17} \\ \hline
\end{tabular}}

\vspace{0.8em}

\subcaptionbox{Recall\label{tab:recall}}{%
\begin{tabular}{ll | cccc | cccc}
\hline
\multirow{2}{*}{\textbf{Training}} & \multirow{2}{*}{\textbf{Method}}
  & \multicolumn{4}{c|}{\textbf{Conflict-AV-MNIST}}
  & \multicolumn{4}{c}{\textbf{Conflict-NYUD}} \\ \cline{3-10}
 & & \textbf{0\%} & \textbf{50\%} & \textbf{75\%} & \textbf{100\%}
   & \textbf{0\%} & \textbf{50\%} & \textbf{75\%} & \textbf{100\%} \\ \hline
\multirow{4}{*}{Decoupled}
 & \cellcolor{cwmgroup}CWM
   & \gradientcell{98.88}{65}{100}{cwmgroupdark}{}
   & \gradientcell{84.20}{65}{100}{cwmgroupdark}{}
   & \gradientcell{76.85}{65}{100}{cwmgroupdark}{}
   & \gradientcell{69.50}{65}{100}{cwmgroupdark}{}
   & \gradientcell{41.71}{35}{65}{cwmgroupdark}{}
   & \gradientcell{41.50}{35}{65}{cwmgroupdark}{}
   & \gradientcell{40.99}{35}{65}{cwmgroupdark}{}
   & \gradientcell{40.82}{35}{65}{cwmgroupdark}{} \\
 & \cellcolor{cwmgroup}\textbf{C$^2$WM}
   & \gradientcell{97.60}{65}{100}{cwmgroupdark}{-1.28}
   & \gradientcell{97.35}{65}{100}{cwmgroupdark}{13.15}
   & \gradientcell{97.22}{65}{100}{cwmgroupdark}{20.37}
   & \gradientcell{97.01}{65}{100}{cwmgroupdark}{27.51}
   & \gradientcell{41.97}{35}{65}{cwmgroupdark}{0.26}
   & \gradientcell{40.69}{35}{65}{cwmgroupdark}{-0.81}
   & \gradientcell{41.54}{35}{65}{cwmgroupdark}{0.55}
   & \gradientcell{41.58}{35}{65}{cwmgroupdark}{0.76} \\ \cline{2-10}
 & \cellcolor{dpcgroup}DPC
   & \gradientcell{98.89}{65}{100}{dpcgroupdark}{}
   & \gradientcell{84.06}{65}{100}{dpcgroupdark}{}
   & \gradientcell{76.78}{65}{100}{dpcgroupdark}{}
   & \gradientcell{69.47}{65}{100}{dpcgroupdark}{}
   & \gradientcell{39.29}{35}{65}{dpcgroupdark}{}
   & \gradientcell{39.53}{35}{65}{dpcgroupdark}{}
   & \gradientcell{39.55}{35}{65}{dpcgroupdark}{}
   & \gradientcell{38.85}{35}{65}{dpcgroupdark}{} \\
 & \cellcolor{dpcgroup}\textbf{C$^2$DPC}
   & \gradientcell{99.11}{65}{100}{dpcgroupdark}{0.22}
   & \gradientcell{99.01}{65}{100}{dpcgroupdark}{14.95}
   & \gradientcell{98.97}{65}{100}{dpcgroupdark}{22.19}
   & \gradientcell{98.92}{65}{100}{dpcgroupdark}{29.45}
   & \gradientcell{49.23}{35}{65}{dpcgroupdark}{9.94}
   & \gradientcell{53.35}{35}{65}{dpcgroupdark}{13.82}
   & \gradientcell{56.60}{35}{65}{dpcgroupdark}{17.05}
   & \gradientcell{60.63}{35}{65}{dpcgroupdark}{21.78} \\ \hline
\multirow{4}{*}{Joint}
 & \cellcolor{cwmgroup}CWM
   & \gradientcell{98.70}{65}{100}{cwmgroupdark}{}
   & \gradientcell{83.93}{65}{100}{cwmgroupdark}{}
   & \gradientcell{76.71}{65}{100}{cwmgroupdark}{}
   & \gradientcell{69.30}{65}{100}{cwmgroupdark}{}
   & \gradientcell{43.07}{35}{60}{cwmgroupdark}{}
   & \gradientcell{42.38}{35}{60}{cwmgroupdark}{}
   & \gradientcell{42.09}{35}{60}{cwmgroupdark}{}
   & \gradientcell{41.95}{35}{60}{cwmgroupdark}{} \\
 & \cellcolor{cwmgroup}\textbf{C$^2$WM}
   & \gradientcell{98.60}{65}{100}{cwmgroupdark}{-0.10}
   & \gradientcell{97.86}{65}{100}{cwmgroupdark}{13.93}
   & \gradientcell{97.64}{65}{100}{cwmgroupdark}{20.93}
   & \gradientcell{97.36}{65}{100}{cwmgroupdark}{28.06}
   & \gradientcell{44.84}{35}{60}{cwmgroupdark}{1.77}
   & \gradientcell{44.13}{35}{60}{cwmgroupdark}{1.75}
   & \gradientcell{43.50}{35}{60}{cwmgroupdark}{1.41}
   & \gradientcell{43.46}{35}{60}{cwmgroupdark}{1.51} \\ \cline{2-10}
 & \cellcolor{dpcgroup}DPC
   & \gradientcell{98.99}{65}{100}{dpcgroupdark}{}
   & \gradientcell{84.21}{65}{100}{dpcgroupdark}{}
   & \gradientcell{76.95}{65}{100}{dpcgroupdark}{}
   & \gradientcell{69.39}{65}{100}{dpcgroupdark}{}
   & \gradientcell{38.49}{35}{60}{dpcgroupdark}{}
   & \gradientcell{40.87}{35}{60}{dpcgroupdark}{}
   & \gradientcell{41.26}{35}{60}{dpcgroupdark}{}
   & \gradientcell{41.96}{35}{60}{dpcgroupdark}{} \\
 & \cellcolor{dpcgroup}\textbf{C$^2$DPC}
   & \gradientcell{99.22}{65}{100}{dpcgroupdark}{0.23}
   & \gradientcell{99.45}{65}{100}{dpcgroupdark}{15.24}
   & \gradientcell{99.52}{65}{100}{dpcgroupdark}{22.57}
   & \gradientcell{99.61}{65}{100}{dpcgroupdark}{30.22}
   & \gradientcell{52.86}{35}{60}{dpcgroupdark}{14.37}
   & \gradientcell{53.31}{35}{60}{dpcgroupdark}{12.44}
   & \gradientcell{53.21}{35}{60}{dpcgroupdark}{11.95}
   & \gradientcell{53.45}{35}{60}{dpcgroupdark}{11.49} \\ \hline
\end{tabular}}

\caption{\textbf{Mean test performance }(\%) at varying levels of cross-modal conflict
$\lambda_\text{test} \in \{0\%, 50\%, 75\%, 100\%\}$ under decoupled and
joint training. \colorbox{cwmgroup}{\vphantom{Ag}Amber} and \colorbox{dpcgroup}{\vphantom{Ag}blue} cells show the CWM-based and DPC-based
comparison respectively. Within each group, darker shades indicate higher performance.
Absolute gain of the context-specific method over its
static baseline in parentheses.}
\label{tab:results}
\end{table*}

\begin{figure*}[t]
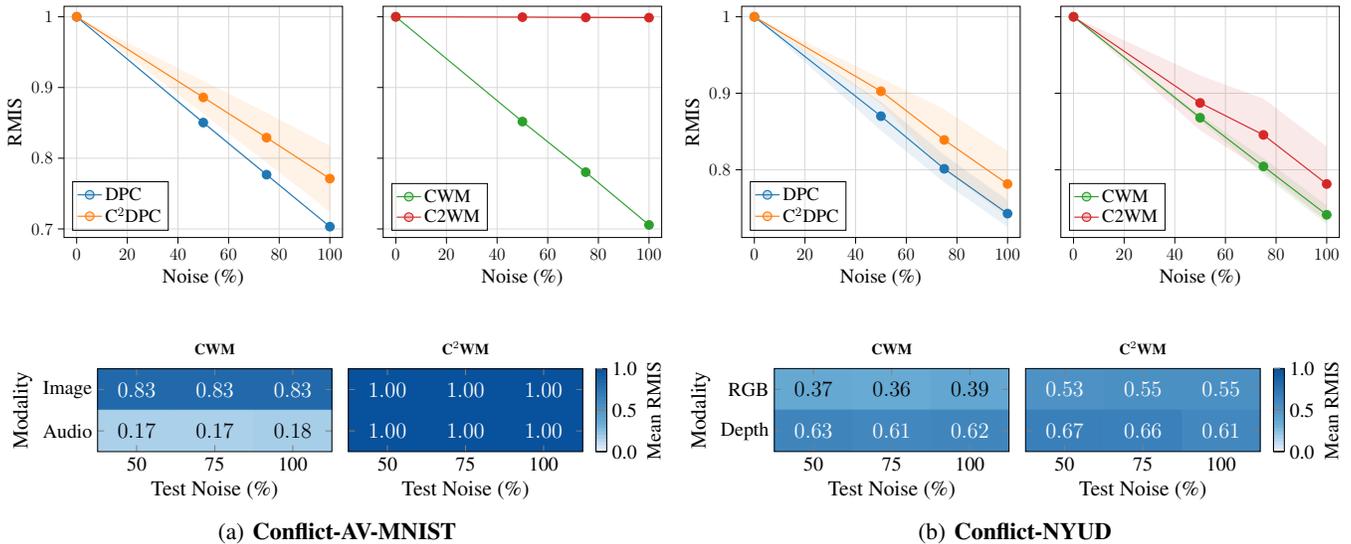

    \centering
    \begin{subfigure}{0.49\textwidth}
        \resizebox{\linewidth}{!}{\begin{tikzpicture}

\definecolor{darkgray176}{RGB}{176,176,176}
\definecolor{darkorange25512714}{RGB}{255,127,14}
\definecolor{steelblue31119180}{RGB}{31,119,180}
\definecolor{forestgreen4416044}{RGB}{44,160,44}
\definecolor{crimson2143940}{RGB}{214,39,40}

\begin{groupplot}[group style={group size=2 by 1}]
\nextgroupplot[
legend cell align={left},
legend style={font=\fontsize{13}{9}\selectfont, fill opacity=0.25, draw opacity=1, text opacity=1, at={(0.03,0.03)}, anchor=south west},
tick align=outside,
tick pos=left,
grid=both,
grid style={line width=0.3pt, draw=gray!30},
xlabel style={font=\fontsize{14}{9}\selectfont},
ylabel style={font=\fontsize{14}{9}\selectfont},
ticklabel style={font=\fontsize{13}{9}\selectfont},
% x grid style={darkgray176},
xlabel={Noise (\%)},
xmin=-5, xmax=105,
xtick style={color=black},
% y grid style={darkgray176},
ylabel={RMIS},
ymin=0.687900019208933, ymax=1.01486190384719,
ytick style={color=black}
]
\path [fill=steelblue31119180, fill opacity=0.1]
(axis cs:0,1)
--(axis cs:0,1)
--(axis cs:50,0.848315123788659)
--(axis cs:75,0.775111934483885)
--(axis cs:100,0.702761923056127)
--(axis cs:100,0.703704719959417)
--(axis cs:100,0.703704719959417)
--(axis cs:75,0.778354743181825)
--(axis cs:50,0.852484857964373)
--(axis cs:0,1)
--cycle;

\path [fill=darkorange25512714, fill opacity=0.1]
(axis cs:0,1)
--(axis cs:0,1)
--(axis cs:50,0.862817756482689)
--(axis cs:75,0.793979954263287)
--(axis cs:100,0.724436405557413)
--(axis cs:100,0.817830241463563)
--(axis cs:100,0.817830241463563)
--(axis cs:75,0.864153353829466)
--(axis cs:50,0.909248916160337)
--(axis cs:0,1)
--cycle;

\addplot [thick, steelblue31119180, mark=*, mark size=3, mark options={solid}]
table {%
0 1
50 0.850399990876516
75 0.776733338832855
100 0.703233321507772
};
\addlegendentry{DPC}
\addplot [thick, darkorange25512714, mark=*, mark size=3, mark options={solid}]
table {%
0 1
50 0.886033336321513
75 0.829066654046377
100 0.771133323510488
};
\addlegendentry{C$^2$DPC}

\nextgroupplot[
legend cell align={left},
legend style={font=\fontsize{13}{9}\selectfont, fill opacity=0.25, draw opacity=1, text opacity=1, at={(0.03,0.03)}, anchor=south west},
scaled y ticks=manual:{}{\pgfmathparse{#1}},
tick align=outside,
tick pos=left,
grid=both,
grid style={line width=0.3pt, draw=gray!30},
xlabel style={font=\fontsize{14}{9}\selectfont},
ylabel style={font=\fontsize{14}{9}\selectfont},
ticklabel style={font=\fontsize{13}{9}\selectfont},
% x grid style={darkgray176},
xlabel={Noise (\%)},
xmin=-5, xmax=105,
xtick style={color=black},
% y grid style={darkgray176},
ymin=0.687900019208933, ymax=1.01486190384719,
ytick style={color=black},
yticklabels={}
]
\path [fill=forestgreen4416044, fill opacity=0.1]
(axis cs:0,1)
--(axis cs:0,1)
--(axis cs:50,0.85124266957378)
--(axis cs:75,0.779309813411357)
--(axis cs:100,0.704579382628783)
--(axis cs:100,0.706753939896241)
--(axis cs:100,0.706753939896241)
--(axis cs:75,0.781223511466065)
--(axis cs:50,0.852423977701509)
--(axis cs:0,1)
--cycle;

\path [fill=crimson2143940, fill opacity=0.1]
(axis cs:0,1)
--(axis cs:0,1)
--(axis cs:50,0.999125843015758)
--(axis cs:75,0.998528467961547)
--(axis cs:100,0.998021104650886)
--(axis cs:100,0.999578894459018)
--(axis cs:100,0.999578894459018)
--(axis cs:75,0.999671551239096)
--(axis cs:50,0.999874149990631)
--(axis cs:0,1)
--cycle;

\addplot [thick, forestgreen4416044, mark=*, mark size=3, mark options={solid}]
table {%
0 1
50 0.851833323637644
75 0.780266662438711
100 0.705666661262512
};
\addlegendentry{CWM}
\addplot [thick, crimson2143940, mark=*, mark size=3, mark options={solid}]
table {%
0 1
50 0.999499996503194
75 0.999100009600321
100 0.998799999554952
};
\addlegendentry{C2WM}
\end{groupplot}

\end{tikzpicture}}
        \label{fig:RMIS_AVMNIST}
    \end{subfigure}
    % \hfill
    \begin{subfigure}{0.49\textwidth}
        \resizebox{\linewidth}{!}{\begin{tikzpicture}

\definecolor{darkgray176}{RGB}{176,176,176}
\definecolor{darkorange25512714}{RGB}{255,127,14}
\definecolor{steelblue31119180}{RGB}{31,119,180}
\definecolor{forestgreen4416044}{RGB}{44,160,44}
\definecolor{crimson2143940}{RGB}{214,39,40}

\begin{groupplot}[group style={group size=2 by 1}]
\nextgroupplot[
legend cell align={left},
legend style={font=\fontsize{13}{9}\selectfont, fill opacity=0.25, draw opacity=1, text opacity=1, at={(0.03,0.03)}, anchor=south west},
tick align=outside,
tick pos=left,
grid=both,
grid style={line width=0.3pt, draw=gray!30},
xlabel style={font=\fontsize{14}{9}\selectfont},
ylabel style={font=\fontsize{14}{9}\selectfont},
ticklabel style={font=\fontsize{13}{9}\selectfont},
% x grid style={darkgray176},
xlabel={Noise (\%)},
xmin=-5, xmax=105,
xtick style={color=black},
% y grid style={darkgray176},
ylabel={RMIS},
ymin=0.711372003393972, ymax=1.01374419031457,
ytick style={color=black}
]
\path [fill=steelblue31119180, fill opacity=0.1]
(axis cs:0,1)
--(axis cs:0,1)
--(axis cs:50,0.851137865527591)
--(axis cs:75,0.783088221798888)
--(axis cs:100,0.725116193708545)
--(axis cs:100,0.760102967007194)
--(axis cs:100,0.760102967007194)
--(axis cs:75,0.81935825481956)
--(axis cs:50,0.888923298374692)
--(axis cs:0,1)
--cycle;

\path [fill=darkorange25512714, fill opacity=0.1]
(axis cs:0,1)
--(axis cs:0,1)
--(axis cs:50,0.885700020991387)
--(axis cs:75,0.798885061294257)
--(axis cs:100,0.738151545058031)
--(axis cs:100,0.824539587169549)
--(axis cs:100,0.824539587169549)
--(axis cs:75,0.878994669566135)
--(axis cs:50,0.919600691594063)
--(axis cs:0,1)
--cycle;

\addplot [thick, steelblue31119180, mark=*, mark size=3, mark options={solid}]
table {%
0 1
50 0.870030581951141
75 0.801223238309224
100 0.74260958035787
};
\addlegendentry{DPC}
\addplot [thick, darkorange25512714, mark=*, mark size=3, mark options={solid}]
table {%
0 1
50 0.902650356292725
75 0.838939865430196
100 0.78134556611379
};
\addlegendentry{C$^2$DPC}

\nextgroupplot[
legend cell align={left},
legend style={font=\fontsize{13}{9}\selectfont, fill opacity=0.25, draw opacity=1, text opacity=1, at={(0.03,0.03)}, anchor=south west},
scaled y ticks=manual:{}{\pgfmathparse{#1}},
tick align=outside,
tick pos=left,
grid=both,
grid style={line width=0.3pt, draw=gray!30},
xlabel style={font=\fontsize{14}{9}\selectfont},
ylabel style={font=\fontsize{14}{9}\selectfont},
ticklabel style={font=\fontsize{13}{9}\selectfont},
% x grid style={darkgray176},
xlabel={Noise (\%)},
xmin=-5, xmax=105,
xtick style={color=black},
% y grid style={darkgray176},
ymin=0.711372003393972, ymax=1.01374419031457,
ytick style={color=black},
yticklabels={}
]
\path [fill=forestgreen4416044, fill opacity=0.1]
(axis cs:0,1)
--(axis cs:0,1)
--(axis cs:50,0.864177732806742)
--(axis cs:75,0.793469338608304)
--(axis cs:100,0.728763473343024)
--(axis cs:100,0.753397611467869)
--(axis cs:100,0.753397611467869)
--(axis cs:75,0.815093369292697)
--(axis cs:50,0.871805956819316)
--(axis cs:0,1)
--cycle;

\path [fill=crimson2143940, fill opacity=0.1]
(axis cs:0,1)
--(axis cs:0,1)
--(axis cs:50,0.851319750222751)
--(axis cs:75,0.798124042181333)
--(axis cs:100,0.732209289233269)
--(axis cs:100,0.830481842994311)
--(axis cs:100,0.830481842994311)
--(axis cs:75,0.893007470142682)
--(axis cs:50,0.923399964895658)
--(axis cs:0,1)
--cycle;

\addplot [thick, forestgreen4416044, mark=*, mark size=3, mark options={solid}]
table {%
0 1
50 0.867991844813029
75 0.8042813539505
100 0.741080542405446
};
\addlegendentry{CWM}
\addplot [thick, crimson2143940, mark=*, mark size=3, mark options={solid}]
table {%
0 1
50 0.887359857559204
75 0.845565756162008
100 0.78134556611379
};
\addlegendentry{C2WM}
\end{groupplot}

\end{tikzpicture}}
        \label{fig:RMIS_NYUD}
    \end{subfigure}

    \vspace{0.5em}

    \begin{subfigure}{0.49\textwidth}
        \resizebox{\linewidth}{!}{\input{CWMvC2WM_avmnist.tex}}
        \caption{\textbf{Conflict-AV-MNIST}}
        \label{fig:AVMNIST_cwm_c2wm}
    \end{subfigure}
    % \hfill
    \begin{subfigure}{0.49\textwidth}
        \resizebox{\linewidth}{!}{\input{CWMvC2WM_nyud.tex}}
        \caption{\textbf{Conflict-NYUD}}
        \label{fig:NYUD_cwm_c2wm}
    \end{subfigure}

    \caption{Mean test RMIS under decoupled training across varying test conflict
    levels $\lambda_\text{test}$. Top row: overall RMIS across all methods.
    Bottom row: RMIS broken down by corrupted modality for CWM vs.\ C$^2$WM,
    revealing whether credibility estimates adapt to the identity of the
    corrupted source or remain biased toward a single modality.}
    \label{fig:rmis_all}
    \vspace{-0.2in}
\end{figure*}
\paragraph{\textbf{Q1: Predictive performance under conflict}}To assess robustness under progressively increasing cross-modal conflict, we set up the following experiments. For each domain, we introduce conflict noise of $\lambda_\text{train} = 0.7$ into the training set. 
% Specifically, for every class designated for image corruption, 70\% of the corresponding image samples are selected and replaced with images drawn from a different target class, thereby inducing cross-modal inconsistency, similarly for the audio modality. This controlled swapping introduces systematic disagreement between modalities.
For the test split, we vary the level of conflict test noise $\lambda_\text{test}$ from 0 (no corruption) to 1.0 (fully corrupted), and evaluate each model at $\lambda \in \{0, 0.5, 0.75, 1.0\}$. 
% This setup enables the assessment of robustness under progressively increasing cross-modal conflict.

We conduct experiments under two distinct training setups:\\
\textbf{(1) Decoupled Training}: In this setting, the unimodal encoders and their corresponding classifiers are first trained independently using their respective unimodal losses. The fusion module is then trained separately, keeping the unimodals frozen, optimizing only the multimodal loss. \\
\textbf{(2) Joint Training}: In this setting, we use the joint training objective $\mathcal{L}_{total}$ and train the entire architecture end-to-end. \\
We present the results from both the training setups here.

Table \ref{tab:results} shows the mean test accuracy, precision, recall, and F1-score, respectively, in both domains. On Conflict-AV-MNIST, the overall test performance decreases as the noise increases. Across all noise settings, our methods C$^2$WM and C$^2$DPC show better performance than their respective baselines. Moreover, this performance gap widens as the $\lambda_\text{test}$ increases which indicates superior robustness of the context-aware fusion methods over their respective baselines. In contrast to Conflict-AV-MNIST, the results from Conflict-NYUD exhibit a less regular trend as $\lambda_\text{test}$ increases. While our methods still outperform their respective baselines, the performance gap doesn't increase as noise increases. We hypothesize that this behaviour arises from the intrinsic noise in the dataset. Increasing the synthetic conflict noise does not necessarily degrade performance, but acts as a regularizer that disrupts the pre-existing modality dominance patterns.

\paragraph{\textbf{Q2: Reliable modality identification}} Figure ~\ref{fig:rmis_all}  reports the mean test RMIS across varying levels of test conflict, $\lambda_\text{test}$. As expected, the mean RMIS value decreases as the test noise increases. Despite this degradation, our methods still consistently achieve higher RMIS values than their respective baselines across all the test noise levels, with C$^2$WM outperforming all the methods. This is because, in C$^2$WM, the computed CSIC values are directly used to weigh the unimodal predictive distributions when computing the multimodal predictions. Since the multimodal predictive loss is obtained from this CSIC-weighted fusion output, the CSIC scores tend to have a strong influence on the multimodal objective. Thus, modalities assigned higher CSIC values contribute more to the objective function. Ideally, as the model converges, the CSIC score associated with the corrupted modality should approach zero, marking the clean modality as the reliable one, thus explaining the consistently better RMIS values for C$^2$WM.

The bottom row of Figure~\ref{fig:rmis_all} further breaks down the mean RMIS scores computed on the corrupted examples by the corrupted modality. 
In Conflict-AV-MNIST, we observe that CWM frequently predicts the image modality as corrupted, even in instances where the audio modality is the one actually corrupted. This misidentification leads to lower RMIS scores, as the model fails to correctly attribute unreliability to the audio modality.
A similar pattern is observed in Conflict-NYUD, where CWM consistently struggles to identify corruption in the RGB modality across all test noise levels.
This behavior suggests that CWM tends to favor one of the modalities as inherently more reliable. This behavior likely arises from its static reliability modelling, which limits its ability to reassess the reliability dynamically.
In contrast, C$^2$WM demonstrates more balanced and accurate identification of the reliable modality across all corrupted modalities and test noise settings. The improved RMIS scores indicate that incorporating additional contextual information enables more adaptive and robust fusion.

\section{Conclusion and future work}
In this work, we addressed the challenge of multimodal fusion 
in environments characterized by 
context-dependent source corruptions. 
% We introduced C$^2$MF, a hybrid neurosymbolic framework that factorizes multimodal integration into a set of neural encoders and predictors combined with Conditional Probabilistic Circuit for late fusion,
% By employing a joint context aggregator $\mathcal{M}$ and a hyper-network $g_\phi$, our approach 
We introduced C$^2$MF, a neurosymbolic framework that combines neural unimodal encoders with a
Conditional Probabilistic Circuit as the fusion function,
satisfying the need for dynamic, instance-specific reliability modeling without sacrificing the rigorous guarantees of probabilistic inference. 
We also introduced the Conflict benchmark and
the RMIS metric, providing the first controlled evaluation framework for studying the efficacy of fusion approaches in cross-modal conflict resolution.
% for evaluating the robustness of fusion approaches under cross-modal conflicts.
Our experimental results 
% on the Conflict-AV-MNIST and Conflict-NYUD datasets 
demonstrated that C$^2$MF significantly outperforms static fusion baselines as cross-modal corruption increases. Beyond predictive performance, our framework also provides a mathematically grounded audit trail through the Context-Specific Information Credibility (CSIC) metric, enabling an exact calculation of each modality's influence on a per-instance basis. This makes it particularly suited for high-stakes domains, such as clinical decision support and autonomous navigation, where sensor reliability is non-stationary.

% This unique combination of robustness and transparency makes C$^2$MF particularly suited for high-stakes domains, such as clinical decision support and autonomous navigation, where sensor reliability is non-stationary.
% While our results on Conflict-AV-MNIST show robustness to class-specific noise, scenarios involving "total signal loss" across all modalities remain an open challenge.
There are several  directions for future work. First, while our benchmarks serve as a diagnostic tool for evaluating context-specific credibility, the corruptions are synthetic, class-aligned and predefined. Real-world conflicts (e.g., sensor degradation due to physical wear) may be more stochastic. Evaluating the framework on large-scale, real-world multimodal datasets with naturally occurring sensor failures, by learning the distribution over corruption masks is an important next step for validating the generalizability of our framework. Second, while our framework achieves better predictive performance than the static PC-based models, it loses the ability to deal with missing data, as C$^2$MF currently requires all modalities to be present for inferring the joint context. Future work should explore ways to adapt C$^2$MF to the missing modality case, such as by replacing the neural models in the context aggregator with a PC \cite{braun2025tractable}. Since C$^2$MF operates well in a decoupled training setting, it can also potentially be extended to combine outputs from multiple pretrained foundation models in a probabilistic manner. Evaluating the context-specific reliability of these models, and fusing the information by weighing their CSIC values is a future direction. Finally, integrating domain knowledge about the relationship between the context and reliability into our framework would further improve performance and remains an important direction for future work.
\vspace{-0.07in}
\section{Acknowledgements}
The authors gratefully acknowledge the support by AFOSR award FA9550-23-1-0239, ARO award W911NF2010224 and DARPA ANSR award HR001122S0039. This work has also benefited from the BMFTR project "Extremely Efficient Inference for Large Context Length (XEI)" (16IS24079B), and also the Cluster of Excellence "Reasonable AI" funded by the German Research Foundation (DFG) under Germany’s Excellence Strategy, EXC-3057.

\bibliographystyle{IEEEtran}
%\bibliography{FUSION2026/references.bib}
\bibliography{refs.bib}
\clearpage

\definecolor{cwmgroup}{RGB}{255, 243, 220}      % warm amber tint (light)
\definecolor{cwmgroupdark}{RGB}{210, 140, 30}   % warm amber dark
\definecolor{dpcgroup}{RGB}{220, 235, 255}      % cool blue tint (light)
% \definecolor{dpcgroupdark}{RGB}{30, 100, 200}   % cool blue dark
\definecolor{dpcgroupdark}{RGB}{125, 180, 255}   % cool blue dark
\pgfplotsset{compat=1.18}
\usepgfplotslibrary{groupplots}
\pgfplotsset{compat=newest}

% % Macro with smaller delta text and automatic contrast switching
% % #1: Value, #2: Min, #3: Max, #4: Color, #5: Delta text
% \newcommand{\gradientcell}[5]{%
%   \edef\percent{\fpeval{round((#1 - #2) / (#3 - #2) * 100, 0)}}%
%   \expanded{\noexpand\cellcolor{#4!\percent!white}}%
%   % \ifnum\percent>70\color{white}\fi% Switch text to white for dark cells
%   #1\ifblank{#5}{}{{\scriptsize\ (#5)}}% Delta in scriptsize for better hierarchy
% }
% \usepackage{mdframed}

\def\BibTeX{{\rm B\kern-.05em{\sc i\kern-.025em b}\kern-.08em
    T\kern-.1667em\lower.7ex\hbox{E}\kern-.125emX}}
% \begin{document}

\onecolumn
\thispagestyle{empty}

\begin{center}
    {\Large \textbf{Supplementary Material}}
\end{center}

\section{Learning Context from Joint Embeddings}
C$2$MF framework uses the context obtained by concatenating the joint embeddings from the unimodal encoders with the corruption mask embeddings. However, the context can be inferred directly from joint unimodal embeddings alone without requiring the corruption masks at inference time. To this end, we trained a Multi-layer Perceptron (MLP) to learn a mapping from the unimodal embeddings to their corresponding corruption masks on the train dataset. At test time, the MLP predicts the context given the joint unimodal embeddings, and this predicted context is mapped to the sum-node parameters of the CPC. Thus, the final predictions of the CPC are dynamically adapted based solely on the inferred context. Table \ref{tab:results_sup} shows the performance of the C$^2$MF methods with this setting for the Conflict-NYUD dataset when trained end-to-end. C$^2$MF methods achieve better performance in terms of all of the metrics over their context-agnostic counterparts.
\begin{table*}[!ht]
\centering
\small
\setlength{\tabcolsep}{8pt}
\renewcommand{\arraystretch}{1.2}

\subcaptionbox{Accuracy\label{tab:accuracy_nyud_joint}}{%
\begin{tabular}{l | cccc}
\hline
\multirow{2}{*}{\textbf{Method}}
  & \multicolumn{4}{c}{\textbf{Test Noise (\%)}} \\ \cline{2-5}
   & \textbf{0\%} & \textbf{50\%} & \textbf{75\%} & \textbf{100\%} \\ \hline

% \multirow{4}{*}{Joint}
 \cellcolor{cwmgroup}CWM
   & \gradientcell{55.15}{45}{65}{cwmgroupdark}{}
   & \gradientcell{54.18}{45}{65}{cwmgroupdark}{}
   & \gradientcell{54.33}{45}{65}{cwmgroupdark}{}
   & \gradientcell{53.98}{45}{65}{cwmgroupdark}{} \\
 \cellcolor{cwmgroup}\textbf{C$^2$WM}
   & \gradientcell{57.29}{45}{65}{cwmgroupdark}{2.14}
   & \gradientcell{56.47}{45}{65}{cwmgroupdark}{2.29}
   & \gradientcell{56.42}{45}{65}{cwmgroupdark}{2.09}
   & \gradientcell{55.56}{45}{65}{cwmgroupdark}{1.55} \\ \cline{2-5}
 \cellcolor{dpcgroup}DPC
   & \gradientcell{50.15}{45}{65}{dpcgroupdark}{}
   & \gradientcell{52.50}{45}{65}{dpcgroupdark}{}
   & \gradientcell{52.70}{45}{65}{dpcgroupdark}{}
   & \gradientcell{53.67}{45}{65}{dpcgroupdark}{} \\
 \cellcolor{dpcgroup}\textbf{C$^2$DPC}
   & \gradientcell{59.12}{45}{65}{dpcgroupdark}{8.97}
   & \gradientcell{57.59}{45}{65}{dpcgroupdark}{5.59}
   & \gradientcell{57.65}{45}{65}{dpcgroupdark}{4.95}
   & \gradientcell{55.71}{45}{65}{dpcgroupdark}{2.04} \\ \hline
\end{tabular}}

\vspace{0.3em}

\subcaptionbox{AUROC\label{tab:auroc_nyud_joint}}{%
\begin{tabular}{l | cccc}
\hline
\multirow{2}{*}{\textbf{Method}}
  & \multicolumn{4}{c}{\textbf{Test Noise (\%)}} \\ \cline{2-5}
   & \textbf{0\%} & \textbf{50\%} & \textbf{75\%} & \textbf{100\%} \\ \hline

 \cellcolor{cwmgroup}CWM
   & \gradientcell{84.65}{80}{100}{cwmgroupdark}{}
   & \gradientcell{85.81}{80}{100}{cwmgroupdark}{}
   & \gradientcell{86.11}{80}{100}{cwmgroupdark}{}
   & \gradientcell{86.27}{80}{100}{cwmgroupdark}{} \\
 \cellcolor{cwmgroup}\textbf{C$^2$WM}
   & \gradientcell{85.70}{80}{100}{cwmgroupdark}{1.05}
   & \gradientcell{85.81}{80}{100}{cwmgroupdark}{0.00}
   & \gradientcell{86.42}{80}{100}{cwmgroupdark}{0.31}
   & \gradientcell{86.81}{80}{100}{cwmgroupdark}{0.54} \\ \cline{2-5}
 \cellcolor{dpcgroup}DPC
   & \gradientcell{82.18}{80}{100}{dpcgroupdark}{}
   & \gradientcell{83.56}{80}{100}{dpcgroupdark}{}
   & \gradientcell{84.36}{80}{100}{dpcgroupdark}{}
   & \gradientcell{85.15}{80}{100}{dpcgroupdark}{} \\
 \cellcolor{dpcgroup}\textbf{C$^2$DPC}
   & \gradientcell{87.25}{80}{100}{dpcgroupdark}{5.07}
   & \gradientcell{87.71}{80}{100}{dpcgroupdark}{4.15}
   & \gradientcell{87.71}{80}{100}{dpcgroupdark}{3.35}
   & \gradientcell{87.38}{80}{100}{dpcgroupdark}{2.23} \\ \hline
\end{tabular}}

\vspace{0.3em}

\subcaptionbox{F1-score\label{tab:f1_nyud_joint}}{%
\begin{tabular}{l | cccc}
\hline
\multirow{2}{*}{\textbf{Method}}
  & \multicolumn{4}{c}{\textbf{Test Noise (\%)}} \\ \cline{2-5}
   & \textbf{0\%} & \textbf{50\%} & \textbf{75\%} & \textbf{100\%} \\ \hline

 \cellcolor{cwmgroup}CWM
   & \gradientcell{42.59}{35}{60}{cwmgroupdark}{}
   & \gradientcell{40.81}{35}{60}{cwmgroupdark}{}
   & \gradientcell{40.09}{35}{60}{cwmgroupdark}{}
   & \gradientcell{39.76}{35}{60}{cwmgroupdark}{} \\
 \cellcolor{cwmgroup}\textbf{C$^2$WM}
   & \gradientcell{43.20}{35}{60}{cwmgroupdark}{0.61}
   & \gradientcell{42.05}{35}{60}{cwmgroupdark}{1.24}
   & \gradientcell{41.92}{35}{60}{cwmgroupdark}{1.83}
   & \gradientcell{40.74}{35}{60}{cwmgroupdark}{0.98} \\ \cline{2-5}
  \cellcolor{dpcgroup}DPC
   & \gradientcell{37.19}{35}{60}{dpcgroupdark}{}
   & \gradientcell{39.10}{35}{60}{dpcgroupdark}{}
   & \gradientcell{39.61}{35}{60}{dpcgroupdark}{}
   & \gradientcell{40.06}{35}{60}{dpcgroupdark}{} \\
  \cellcolor{dpcgroup}\textbf{C$^2$DPC}
   & \gradientcell{42.85}{35}{60}{dpcgroupdark}{5.66}
   & \gradientcell{48.85}{35}{60}{dpcgroupdark}{9.75}
   & \gradientcell{48.62}{35}{60}{dpcgroupdark}{9.01}
   & \gradientcell{45.52}{35}{60}{dpcgroupdark}{5.46} \\ \hline
\end{tabular}}

\vspace{0.3em}

\subcaptionbox{Precicion\label{tab:precision_nyud_joint}}{%
\begin{tabular}{l | cccc}
\hline
\multirow{2}{*}{\textbf{Method}}
  & \multicolumn{4}{c}{\textbf{Test Noise (\%)}} \\ \cline{2-5}
   & \textbf{0\%} & \textbf{50\%} & \textbf{75\%} & \textbf{100\%} \\ \hline

% \multirow{4}{*}{Joint}
 \cellcolor{cwmgroup}CWM
   & \gradientcell{46.79}{35}{60}{cwmgroupdark}{}
   & \gradientcell{41.42}{35}{60}{cwmgroupdark}{}
   & \gradientcell{39.15}{35}{60}{cwmgroupdark}{}
   & \gradientcell{38.96}{35}{60}{cwmgroupdark}{} \\
  \cellcolor{cwmgroup}\textbf{C$^2$WM}
   & \gradientcell{43.02}{35}{60}{cwmgroupdark}{-3.77}
   & \gradientcell{41.27}{35}{60}{cwmgroupdark}{-0.15}
   & \gradientcell{41.39}{35}{60}{cwmgroupdark}{2.24}
   & \gradientcell{40.49}{35}{60}{cwmgroupdark}{1.53} \\ \cline{2-5}
  \cellcolor{dpcgroup}DPC
   & \gradientcell{41.96}{35}{60}{dpcgroupdark}{}
   & \gradientcell{46.96}{35}{60}{dpcgroupdark}{}
   & \gradientcell{45.94}{35}{60}{dpcgroupdark}{}
   & \gradientcell{46.77}{35}{60}{dpcgroupdark}{} \\
  \cellcolor{dpcgroup}\textbf{C$^2$DPC}
   & \gradientcell{55.03}{35}{60}{dpcgroupdark}{13.07}
   & \gradientcell{50.89}{35}{60}{dpcgroupdark}{3.93}
   & \gradientcell{50.79}{35}{60}{dpcgroupdark}{4.85}
   & \gradientcell{48.34}{35}{60}{dpcgroupdark}{1.57} \\ \hline
\end{tabular}}

\vspace{0.3em}

\subcaptionbox{Recall\label{tab:recall_nyud_joint}}{%
\begin{tabular}{l | cccc}
\hline
\multirow{2}{*}{\textbf{Method}}
  & \multicolumn{4}{c}{\textbf{Test Noise (\%)}} \\ \cline{2-5}
   & \textbf{0\%} & \textbf{50\%} & \textbf{75\%} & \textbf{100\%} \\ \hline

% \multirow{4}{*}{Joint}
 \cellcolor{cwmgroup}CWM
   & \gradientcell{43.07}{35}{60}{cwmgroupdark}{}
   & \gradientcell{42.38}{35}{60}{cwmgroupdark}{}
   & \gradientcell{42.09}{35}{60}{cwmgroupdark}{}
   & \gradientcell{41.95}{35}{60}{cwmgroupdark}{} \\
  \cellcolor{cwmgroup}\textbf{C$^2$WM}
   & \gradientcell{44.11}{35}{60}{cwmgroupdark}{1.04}
   & \gradientcell{43.74}{35}{60}{cwmgroupdark}{1.36}
   & \gradientcell{43.71}{35}{60}{cwmgroupdark}{1.62}
   & \gradientcell{42.77}{35}{60}{cwmgroupdark}{0.82} \\ \cline{2-5}
  \cellcolor{dpcgroup}DPC
   & \gradientcell{38.49}{35}{60}{dpcgroupdark}{}
   & \gradientcell{40.87}{35}{60}{dpcgroupdark}{}
   & \gradientcell{41.26}{35}{60}{dpcgroupdark}{}
   & \gradientcell{41.96}{35}{60}{dpcgroupdark}{} \\
  \cellcolor{dpcgroup}\textbf{C$^2$DPC}
   & \gradientcell{52.12}{35}{60}{dpcgroupdark}{13.63}
   & \gradientcell{48.42}{35}{60}{dpcgroupdark}{7.55}
   & \gradientcell{48.38}{35}{60}{dpcgroupdark}{7.12}
   & \gradientcell{45.30}{35}{60}{dpcgroupdark}{3.34} \\ \hline
\end{tabular}}

\caption{\textbf{Mean test performance }(\%) at varying levels of cross-modal conflict
$\lambda_\text{test} \in \{0\%, 50\%, 75\%, 100\%\}$ under joint training for the Conflict-NYUD dataset. \colorbox{cwmgroup}{\vphantom{Ag}Amber} and \colorbox{dpcgroup}{\vphantom{Ag}blue} cells show the CWM-based and DPC-based comparison respectively. Within each group, darker shades indicate higher performance. Absolute gain of the context-specific method over its static baseline in parentheses.}
\label{tab:results_sup}
\vspace{-0.2in}
\end{table*}

% \end{document}

% \input{FUSION2026/notations}
\end{document}